\documentclass[smallextended]{svjour3}       % onecolumn (second format)
\usepackage{graphicx}
\graphicspath{{resource/}}
\usepackage{subfigure}
\usepackage{multirow}
\usepackage{multicol} 
\usepackage{makecell}
\usepackage{color}
\usepackage{algorithm}
\usepackage{algorithmic}
\usepackage{mathrsfs}
\usepackage{amsmath}
\usepackage{amssymb}

\begin{document}

\title{Learning Discriminative Hashing Codes for Cross-Modal Retrieval based on Multi-view Features%\thanks{Grants or other notes
}


\author{
	Jun Yu \and
        Xiao-Jun Wu \and Josef Kittler
}

%\institute{
%}
%\institute{J. Yu \at
%              the School of Internet of Things Engineering, Jiangnan University, Wuxi 214122, China \\
%%              Tel.: +123-45-678910\\
%%              Fax: +123-45-678910\\
%              \email{yujunjason@aliyun.com}           %  \\
%%             \emph{Present address:} of F. Author  %  if needed
%          X.-J. Wu (Corresponding Author)\at
%              the School of Internet of Things Engineering, Jiangnan University, Wuxi 214122, China\\
%		\email{wu\_xiaojun@jiangnan.edu.cn}
%		\and
%	       J. Kittler\at
%             the Centre for Vision, Speech and Signal Processing, University of Surrey, Guildford GU2 7XH, U.K.
%		\email{j.kittler@surrey.ac.uk}
%}

\date{}

\maketitle

\begin{abstract}
Hashing techniques have been applied broadly in retrieval tasks due to their low storage requirements and high speed of processing. Many hashing methods based on a single view have been extensively studied for information retrieval. However, the representation capacity of a single view is insufficient and some discriminative information is not captured, which results in limited improvement. In this paper, we employ multiple views to represent images and texts for enriching the feature information. Our framework exploits the complementary information among multiple views to better learn the discriminative compact hash codes. A discrete hashing learning framework that jointly performs classifier learning and subspace learning is proposed to complete multiple search tasks simultaneously. Our framework includes two stages, namely a kernelization process and a quantization process. Kernelization aims to find a common subspace where multi-view features can be fused. The quantization stage is designed to learn discriminative unified hashing codes. Extensive experiments are performed on single-label datasets (WiKi and MMED) and multi-label datasets (MIRFlickr and NUS-WIDE) and the experimental results indicate the superiority of our method compared with the state-of-the-art methods.
\keywords{cross-modal retrieval\and hashing learning\and subspace learning \and kernelization \and multi-view features. }
\end{abstract}

\section{INTRODUCTION}
\hspace*{0.5cm}Recently, the explosive growth of multimedia data on the Internet has magnified the challenge of information retrieval. Multimedia data usually emerges in different forms, such as image, text, video, and audio. Hashing, is an effective feature representation of data, the purpose of which is to enhance the retrieval efficiency. It has received considerable attention in multimedia data analysis and retrieval applications\cite{1}.Unimodal retrieval, which implements a query within a single data type, is extensively applied in practice. For example, an image query specified by a given image returns relevant images from the database. Many unimodal methods, such as Semi-supervised Hashing (SSH) \cite{2}, Semi-supervised manifold-embedded hashing (SMH) \cite{42}, Minimal Loss Hashing (MLH) \cite{3} based on the latent structural SVM framework, learn the hashing mapping by employing labeled data. Weakly Supervised Multimodal Hashing (WMH)\cite{paa7} integrates the weakly supervised tag information, underlying discriminative information and the geometric structure in the visual space to learn semantic-aware hash functions for scalable social image retrieval. Besides the supervised approaches, some unsupervised methods, such as Scalable Graph Hashing with feature transformation (SGH)\cite{4}, Anchor Graph-based Hashing (AGH) \cite{5} and Contour Points Distribution Histogram (CPDH)\cite{7} also achieve high performance.
However, unimodal retrieval methods are not directly extensive to multimedia data. This has motivated the development of cross-modal hashing methods \cite{8,9,10,11,12,13,14,15,6,16} which support searching among multi-modal data. The most popular Canonical Correlation Analysis (CCA) \cite{23} projects multiple modal data into a common latent subspace where the correlation among projected vectors of all modalities is maximized. Semi-paired Hashing \cite{24} learns semi-supervised hashing by jointly performing feature extraction and classifier learning. However, These methods do not obtain satisfying retrieval precision because the learned hashing codes lack discriminative ability. In contrast, most of the supervised cross modal methods \cite{15,6,17,18,19,20} which use the semantic label for hash code learning have achieved promising performance. An example is Semantic Correlation Maximization (SCM) \cite{15} which integrates semantic labels into the learning framework. However, the above methods just use single view to capture feature information. The representation ability of single view feature is limited and the inner structural information of an object is not captured effectively in the above methods. Yet, the structural information is very important to characterize the correlations among features. It is therefore desirable to explore as much structural information converged by feature spaces as possible. The covariance matrix \cite{27},which is considered as the representation of the second order data statistics, captures the structural information of the feature space. The histogram feature and the first-order feature further enhance the feature representation ability. Inspired by the Multi-kernel Metric Learning method \cite{25,26}, we propose a novel learning framework, termed Multi-view Feature Discrete Hashing (MFDH), which exploits the complementary information provided by multiple views information and combines the classifier learning to learn the discriminative unified hashing code.
The flowchart of the framework of the proposed MFDH is illustrated in Fig.1.
The main contributions of the proposed MFDH are: \\\
1) A kernel approach is used to learn a common subspace in which multi-view features located in heterogeneous spaces are fused in order to enhance the representation capacity and enrich the feature information.\\
2) The multi-view information, semantic label information are incorporated into a unified framework for multiple retrieval tasks under different data scenarios.\\
3) A comparative evaluation of the proposed method on four available datasets with other state-of-the-art hashing methods, which shows that MFDH boosts the retrieval performance.\\
\hspace*{0.5cm}Structurally, the rest of this paper falls into five parts. In Section 2, we review related works about Cross-modal retrieval. The structure of MFDH model is described in Section 3, and the joint optimization process of MFDH is presented in Section 4. The experimental results and analysis are depicted in Section 5, and finally we draw the conclusions of the paper in Section 6.
%however, the inner structural information of an object is not utilized directly in the process of vectorization of data matrices. Yet, the structural information is very important to characterize the correlations among features. It is therefore desirable to exploit as much structural information converged by feature spaces as possible. Inspired by the Multi-kernel Metric Learning method \cite{25,26}, we propose a novel learning framework, termed Multi-view Feature Discrete Hashing (MFDH), which fuses multiple view features to learn a discriminative unified hashing code. It not only integrates the complementary information provided by multiple modal data but also enriches the structure information of samples of each modality. The covariance matrix \cite{27},which is considered as the representation of the second order data statistics, captures the structural information of the feature space. The histogram feature and the first order statistical feature further enrich the representation. Since the histogram feature, the first order feature and the second order feature belong to different Euclidean spaces and Riemannian manifolds respectively, weintroduce a multi-kernel method to learn a common subspace where multi-view statistical features are fused.
%Composite Correlation Quantization (CCQ) is a cross-modal quantization method which learns the sparse representation by introducing a dictionary matrix constructed by multiple small dictionarys.
\section{RELATED WORK}
\hspace*{0.5cm}Cross-modal retrieval has been a popular topic in multimedia applications. In this section, we preliminarily review the related work of cross-modal retrieval. According to the different ways of feature representations, cross-modal retrieval methods are divided into two branches: real-valued representation learning \cite{21,22,paa1,paa8,paa2,paa3,27} and hash representation learning \cite{23,24,r1,r2,r3,r4,r5,r6,r7,r8,28,39,40,41}. Real-valued representation learning is to find a common subspace where the different media data can be directly compared to each other. Deep Collaborative Embedding (DCE)\cite{paa1} which employs the weakly supervised information, visual structure information and tag correlation information to learn a latent subspace is the first attempt to address the cross-modal search, Content-based Image Retrieval(CBIR) and tag expansion simultaneously under the deep factor analysis framework. Yu et al. \cite{paa8} proposed to uncover a common subspace by maximizing kernel correlation and preserving discriminative structure. Yang et al. \cite{paa2} proposed to learn shared semantic space by aligning the correlation of two modalities.\\
\hspace*{0.5cm}The approach proposed in this paper falls in the branch of hash representation learning. The hash representation learning is also refferred to as Cross-modal hashing which are more geared towards retrieval efficiency. Cross-modal hashing usually aims to learn a shared Hamming space where the distance among multiple modalities can be measured. Various hashing models have been proposed for efficient cross-modal retrieval. Cross-modal hashing methods are roughly grouped into two categories including Unsupervised Cross-modal Hashing and Supervised Cross-modal Hashing.\\
\hspace*{0.5cm}Unsupervised cross-modal hashing \cite{23,r1,r2,r3,r4,r5,r6} mainly learn the hash functions by exploring the structural, distribution, topological information of data. Canonical Correlation Analysis (CCA) \cite{23} learns a common space where the correlation between different two modalities is maximized. Inter-Media Hashing (IMH) \cite{r1} preserves both inter-media and intra-media consistency to learn their binary features. Cross View Hashing (CVH) \cite{r2} extends the classical unimodal spectral hashing to the multi-modal scenario. Robust Cross-view Hashing(RCH) \cite{r3} learns a common Hamming space where the binary codes presented different modalities but the same semantic content are as consistent as possible. Collective Reconstructive Embeddings (CRE) \cite{r4} directly learns the unified binary codes via reconstructive embeddings collectively. Fusion Similarity Hashing(FSH) \cite{r5} constructs an undirected asymmetric graph to model the similarity among different modalities. Robust and Flexible Discrete Hashing(RFDH) \cite{r6} adopts the discrete matrix decomposition to learn the binary codes, which avoids the large quantization error caused by relaxation\\
\hspace*{0.5cm}Different from unsupervised cross-modal hashing, supervised cross-modal hashing methods \cite{15,r7,r8,39,40} exploit the available semantic label information to learn the discriminative hashing feature. Semantic Correlation Maximization (SCM) \cite{15} preserves the cosine similarity calculated by the label vectors in Hamming space. Supervised Matrix Factorization Hashing (SMFH) \cite{r7} integrates the graph regularization and matrix factorization into the hashing learning framework. Semantics-Preserving Hashing (SePH) \cite{40} transforms the affinity matrix into a probability distribution and the distribution is preserved in Hamming space via minimizing their Kullback-Leibler divergence. Generalized Semantic Preserving Hashing (GSePh) \cite{r8} preserves the semantic similarity between multi-modal data to learn the unified binary codes for  multiple scenarios. Discrete Cross-modal Hashing (DCH) \cite{39} retains the discrete constraints to learn the discriminative binary codes. To better capture the nonlinear structural information underlying the original features, DCH+RBF extends the DCH to nonlinear embedding by using the typical Gaussian kernel. Above methods have shown promising performance but there are still shortcomings. Since the diversity of the data structure and distribution for different original data, single view feature is not enough to represent the feature of all samples, our model introduces multi-view features into the hashing learning framework to learn the discriminative hashing codes.
\begin{figure}
  \includegraphics[width=\textwidth]{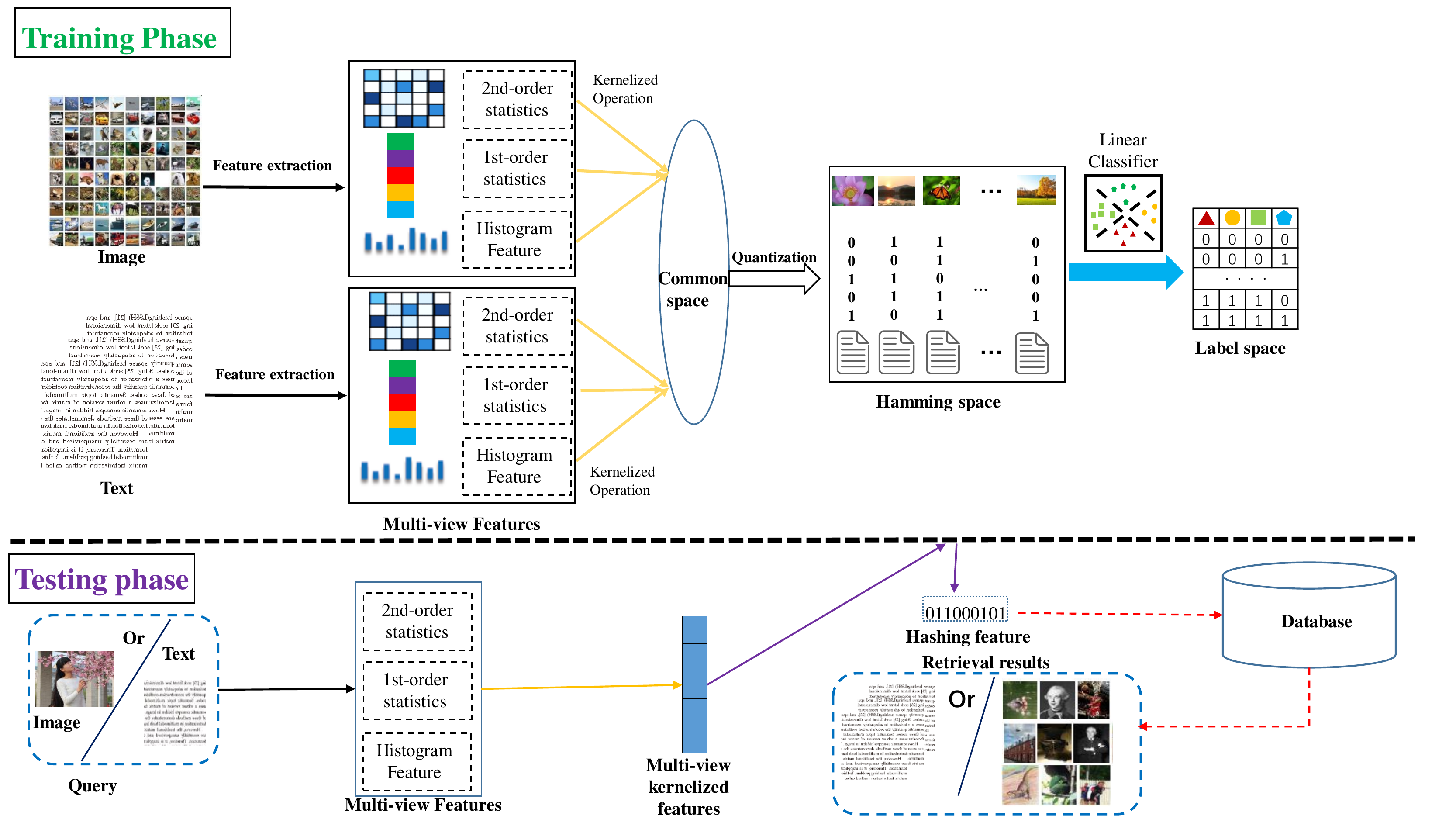}
\caption{The MFDH flowchart. In the training phase, MFDH consists of four steps: Firstly, we extract the multi-view features of the original multi-modal data (image or text). Secondly, kernelized multi-view features are projected to a common subspace. Thirdly, the fused features in the common space are quantized into binary codes. Finally, samples represented by the hashing feature are mapped into the Label space. In the testing phase, we first obtain the multi-view features of arbitrary query data (an image or a text) and calculate the kernelized feature of the query. Then the hashing feature can be got according to the learned hashing functions. After the XOR operation, another modal data closing to the query data is returned from the database.}
\end{figure}

\section{PROPOSED METHOD}
\subsection{Sample Representation}
\hspace*{0.5cm}Suppose that there are $n$ instances $O=\{o_1, o_2, o_3...o_n\}$ of image-text pairs in the training set. For each instance $o_i = \{v_i,t_i\}$, $v_i$ is the fused feature of the image and $t_i$ is the fused feature of the text. $I=\{v_1, v_2, v_3...v_n\}$ and $T=\{t_1, t_2, t_3...t_n\}$ signify image-modality and text-modality respectively. The classification label matrix is denoted by $Y = [y_1, y_2, y_3... y_n] \in R^{c\times n}$, where $c$ is the number of categories.
\subsection{Multi-view Features}
\hspace*{0.5cm}In many traditional approaches, every sample, i.e. two-dimension image or text, is usually transformed into one-dimension feature vector. In this process, the structural information of the data is frequently ignored. However, the structural information conveys important information content. To exploit the underlying structural information, multi-view statistics of the two modalities are adopted. Multi-view features are presented as follows:\\
1) \begin{bfseries}Histogram feature:\end{bfseries} MFDH divides all images into many patches with constant size, and computes local descriptors by employing dense SIFT operator \cite{29,30,31}. Thus, every image is modeled by a set of local descriptors. Specifically, the $i-th$ image is represented as $G_{ij}^{img}$, $j=1,…,N_i^{img}$, where $N_i^{img}$ denotes the number of patches in the $i-th$ image. Then, we learn a dictionary whose size is $k$ for all local feature descriptors of all images based on the k-means clustering algorithm. After that, each image is quantized into a histogram feature vector according to the Bag Of View Word (BOVW) model \cite{32}. Specifically, the histogram feature of the $i-th$ image is denoted as $z_i^{img} (i=1,2,…,n)$. For text, we use word vectors learned according to the word2vec model \cite{33} to learn a dictionary by adopting k-means algorithm. Likewise, the high-dimensional histogram feature vector $z_i^{txt} (i=1,2,…,n)$ of the $i-th$ text can be obtained. The above histogram feature vector is regarded as \begin{bfseries}zeroth-order feature\end{bfseries} in this section.\\
2) \begin{bfseries}First-order feature:\end{bfseries} For the $i-th$ image, we compute the mean vector $m_i^{img}$ as the first-order feature.
\begin{equation}
m_i^{img}=\frac{1}{N_i^{img}}\sum_{j=1}^{N_i^{img}}G_{ij}^{img}    
\end{equation}
Similarly, we can also get the mean vector $m_i^{txt}$ for the $i-th$ text.\\
3) \begin{bfseries}Second-order feature:\end{bfseries} We select the feature representation based on covariance matrix as the second-order statistical feature of samples. The diagonal element of covariance matrix is the variance of individual components of the representation vector, and non-diagonal elements reflect the correlations between the different components. There are two main advantages of the feature representation based on covariance matrix \cite{27,34}. On the one hand, the covariance encodes the feature correlation information of each class to better discriminate the samples from different categories. Specifically, the local descriptors from different two samples belonging to the same category should be close to each other in high-dimension feature space, since they encode the same semantic content. Thus, the corresponding entries of their covariance matrix should also be close enough to each other. On the other hand, the covariance matrix-based representation can effectively filter out local feature descriptors corrupted by noise within sample because of the averaging process in computing the covariance. The second-order statistics of the $i-th$ image is obtained as 
\begin{equation}
C_i^{img}=\frac{1}{N_i^{img}-1}\sum_{j=1}^{N_i^{img}}(G_{ij}^{img}-m_i^{img})(G_{ij}^{img}-m_i^{img})^T   
\end{equation}
In the same way, we also compute the covariance matrix based representation for text.\\
\hspace*{0.5cm}After computing multi-view features, we represent the $i-th$ image with triplet $(z_i^{img},m_i^{img},C_i^{img})$ and the $i-th$ text is denoted as $(z_i^{txt},m_i^{txt},C_i^{txt})$.\\
\begin{figure*}
\centering
  \includegraphics[width=.7\textwidth]{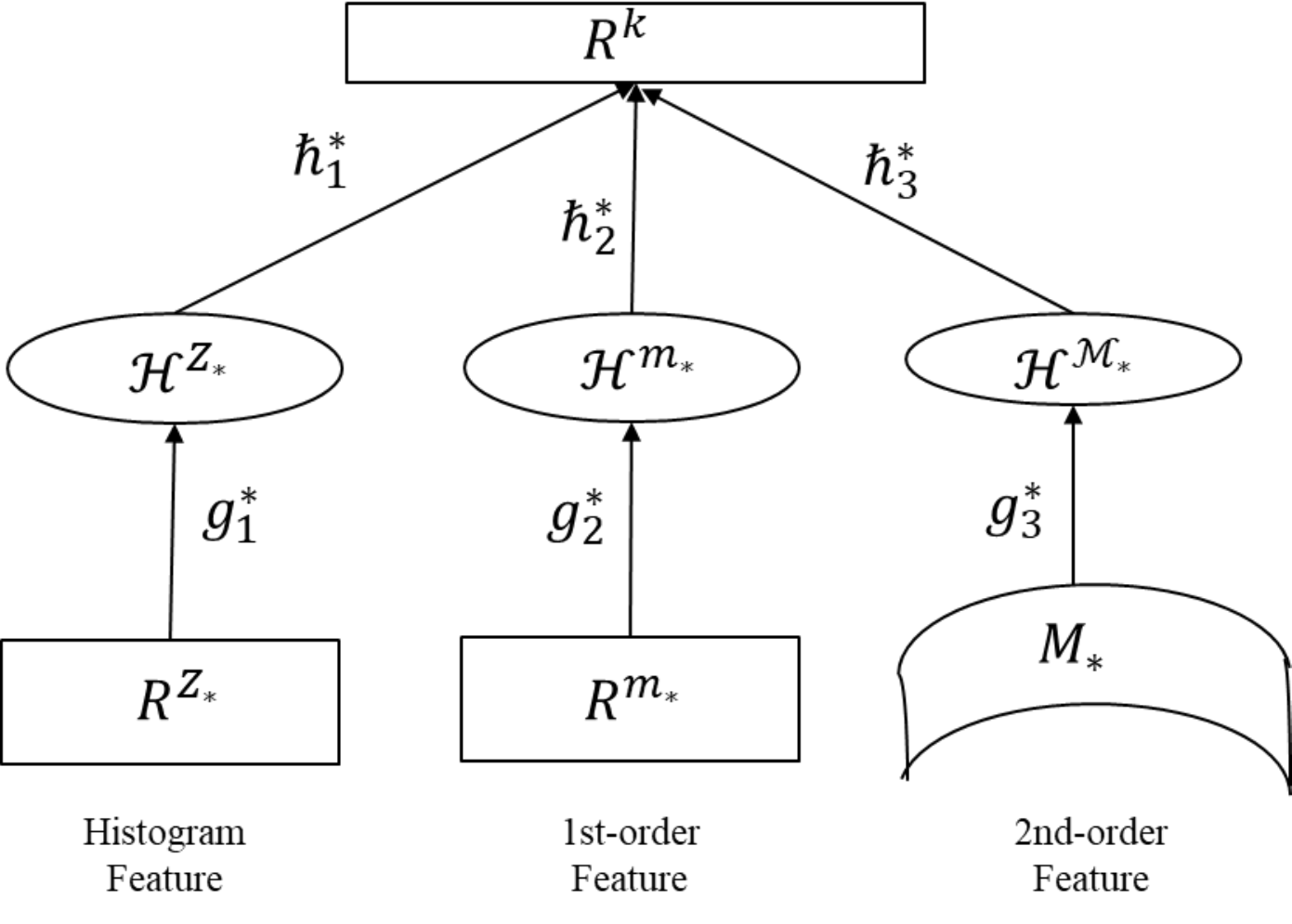}
\caption{The mapping modes of the kernel operation. The model represents the kernelization process of multi-view features of image or text respectively ('*' denotes 'I' or 'T')}
\end{figure*}
\subsection{Kernelized Operation}
\hspace*{0.5cm}The points represented by $z_i^{img}$, $m_i^{img}$ and $z_i^{txt}$, $m_i^{txt}$ are located in Euclidean spaces $R^{Z_I}$,$R^{m_I}$ and $R^{Z_T}$,$R^{m_T}$ respectively, while the second-order statistics represented by covariance matrices $C_i^{img}$, $C_i^{txt}$, which approximately lie on the Riemannian manifolds $\mathcal{M}_I$, $\mathcal{M}_T$ respectively. It is difficult to fuse multi-view features directly since they locate in heterogeneous spaces. To tackle this problem, we use the kernel trick to embed the heterogeneous spaces into high-dimensional kernel spaces. As shown in Fig.2, the model includes implicit mapping $g_r^*(r = 1, 2, 3)^1$ that maps the original data represented by the $(r-1)-th$ order feature to Hilbert space $\mathcal{H}$ and the transformation functions $\hslash_r^*$ mapping the data in Hilbert space $\mathcal{H}$ to common space $R^k$ ('*' signifies image-modality or text-modality when it is 'img'or 'txt' in this paper.). Here we use $x_{ir}^*$ to represent the $r-th$ term of the triplet $(z_i^*,m_i^*,C_i^*)$. The final transformation function $\mathcal{F}_r^{img}= \hslash_r^{img}\circ g_r^{img}$ for image, where the linear projections $\hslash_r^{img} (x_{ir}^{img})=w_r^{img} [g_r^{img}(x_{ir}^{img})]$. According to the kernel trick of Support Vector Machine(SVM), the function $\mathcal{F}_r^{img}$ can be transformed as $\mathcal{F}_r^{img}(x_{ir}^{img})=w_r^{img}[g_r^{img}(x_{ir}^{img})]=\sum_j^{d_r}p_{jr}^{img}[g_r^{img}(x_{jr}^{img})]^T[g_r^{img}(x_{ir}^{img})]=p_r^{img}K_{ir}^{img}$, where $w_r^{img}$ denotes a latent projection matrix that maps the $r-th$ view feature of the image-modality from the Hilbert space to the common space and it is approximated by the linear combination of $d_r$ samples selected in Hilbert space and the $j-th$ element of $K_{ir}^{img}$ is $<g_r^{img}(x_{ir}^{img}),g_r^{img}(x_{jr}^{img})>$. Likewise, we can obtain the function $\mathcal{F}_r^{txt}(x_{ir}^{txt})=p_r^{txt}K_{ir}^{txt}$ of the $(r-1)-th$ order feature for text-modality. We introduce RBF Kernel function $k(i,j)$ in Eq. (3) and polynomial kernel function $\tilde{k} (i,j)$in Eq. (4).
\begin{equation}
k(i,j)=exp(-d^2_{x^*_{ir},x^*_{jr}}/2\sigma^2)
\end{equation}
\begin{equation}
\tilde{k}(i,j)=(x^*_{ir}x^*_{jr}+a)^s
\end{equation}
where $a$ and $s$ are coefficients. \\
\hspace*{0.5cm}The distance $d_{x_{ir}^*,x_{jr}^*}^2$ between any two points $x_{ir}^*$, $x_{jr}^*$ which lie on the manifold is measured by the Log-Euclidean Distance (LED).  For the same order features of two modalities (i.e. image and text), we select the same kernel function. $\varphi^{(r)}=[K_{1r}^{img},K_{2r}^{img},...,K_{nr}^{img}]\in R^{d_r\times n}$ and $\phi^{(r)}=[K_{1r}^{txt},K_{2r}^{txt},...,K_{nr}^{txt}]\in R^{d_r\times n}$ denote the kernel matrices of the$(r-1)-th$ order features of training images and texts respectively. The multi-view kernelized features of the $i-th$ image and text are represented as $\{K_{i1}^{img}, K_{i2}^{img}, K_{i3}^{img}\}$ and $\{K_{i1}^{txt}, K_{i2}^{txt}, K_{i3}^{txt}\}$ respectively. The above procedure of acquiring fusion features is called as kernelization process.\\
\subsection{Learning Discriminative Hashing codes}
\hspace*{0.5cm}To fuse multi-view kernelized features, two projection matrices needed to be learned are denoted as $P_{img}=(p_1^{img},p_2^{img},p_3^{img})\in R^{L\times \sum_{r=1}^3d_r}$ and $P_{txt}=(p_1^{txt},p_2^{txt},p_3^{txt})\in R^{L\times \sum_{r=1}^3d_r}$ that transform multi-view kernelized features of image modality and text modality respectively into a common subspace, where $L$ is the length of hash code. $v_i=\sum_{r=1}^3\eta_r^{img}p_r^{img}K_{ir}^{img}=P_{img}\Psi_i$ denotes the fused features of the $i-th$ image, where $\Psi_i$ is the $i-th$ column of $\Psi=(\varphi^{(1)^T},\varphi^{(2)^T},\varphi^{(3)^T})^T$ and $\eta_r^{img}$ is a weighting coefficient. The fused features of the $i-th$ text is represented by $t_i=\sum_{r=1}^3\eta_r^{txt}p_r^{txt}K_{ir}^{txt}=P_{txt}\Phi_i$, where $\Phi_i$ is the $i-th$ column of $\Phi=(\phi^{(1)^T},\phi^{(2)^T},\phi^{(3)^T})^T$ and $\eta_r^{txt}$ is weighting coefficient. In MFDH, an instance $o_i$  is represented jointly by $v_i$ and $t_i$. Since the two modalities have the same semantic information, $v_i$ and $t_i$ can be quantized into the same hashing feature $b_i$ in the hamming space. Concretely, the process can be defined as $v_i \xrightarrow{f} b_i,t_i \xrightarrow{f} b_i$, where $f$ is the sign function  
\begin{equation}
sign(x)=
\begin{cases}
-1& x<0\\
\quad 1& x \ge 0
\end{cases}
\end{equation}
\hspace*{0.5cm}We expect that the learned hash codes are discriminative enough and the class information of samples can be predicted effectively by a linear classifier $W\in R^{L \times c}$.\\
\begin{equation}
y_i=W^Tb_i,i=1,2,…,n.
\end{equation}
We formulate the quantization process of our framework by minimizing the error both projection and classification as follows:
\begin{equation}
\begin{split}
\min_{W,P_{img},P_{txt},B}&\sum_{i=1}^n\|y_i-W^Tb_i\|_F^2+\alpha\sum_{i=1}^n\| b_i-P_{img}\Psi_i\|_F^2\\&+\beta\sum_{i=1}^n\|b_i-P_{txt}\Phi_i\|_F^2 +\lambda\|W\|_F^2,\\&s.t.  b_i\in \{-1,1\}^L
\end{split}
\end{equation}
where $\alpha,\beta$ are the penalty parameters, and $\lambda$ is the regularization parameter.\\
\hspace*{0.5cm}In Equation (7), the fourth term and the first term are an unified term whose role is supervised classification, and the second and third terms aim to obtain the binary codes with less error for image modality and text modality respectively.

\section{OPTIMIZATION ALGORITHM}
\hspace*{0.5cm}The objective function (7) is rewritten as (8) in order to make the optimization procedure more intuitive.\\
\begin{equation}
\underset{B,W,P_I,P_T}{\textrm{min}} \left \|Y-W^TB \right \|_F^2+\alpha \left \|B-P_{img}\Psi \right \|_F^2\\+\beta \left \|B-P_{txt}\Phi \right \|_F^2+\lambda \left \|W \right \|_F^2\\s.t.B\in \{-1,1\}^{L\times n}
\end{equation}
where the each column of $B$ represents an instance. We adopt the ADMM optimization algorithm to solve the problem in (8) with binary constraint. The error is large if relaxing the constraints of $B$ to be continuous, thus we introduce the discrete cyclic coordinate descent method (DCC) \cite{38} to optimize matrix $B$ directly.\\ 
1) \begin{bfseries}Optimization of $P_{img}$:\end{bfseries} Keeping only the terms relating to $P_{img}$, we have

\begin{equation}
\min_{P_{img}}\alpha \left \|B-P_{img}\Psi \right \|_F^2
\end{equation}

A closed-form solution of E.q.9 can be obtained as \\
\begin{equation}
P_{img}=B\Psi^T(\Psi\Psi^T)^{-1}
\end{equation}
2) \begin{bfseries}Optimization of $P_{txt}$:\end{bfseries} Keeping only the terms relating to $P_{txt}$. Similarly, the optimal solution with respect to $P_{txt}$ is
\begin{equation}
P_{txt}=B\Phi^T(\Phi\Phi^T)^{-1} 
\end{equation}
3) \begin{bfseries}Optimization of $W$:\end{bfseries} With all the variables but $W$ fixed, problem (8) can be rewritten as

\begin{equation}
\underset{W}{\textrm{min}}\left \|Y-W^TB \right \|_F^2+\lambda \left \|W \right \|_F^2
\end{equation}
%\min_W\left \|Y-W^TB \right \|_F^2+\lambda \left \|W \right \|_F^2

Setting the derivative of (12) with respect to $W$ to zero, we can get the optimal $W$ as
\begin{equation}
W=(BB^T+\lambda I)^{-1}BY^T
\end{equation}
4) \begin{bfseries}Optimization of $B$:\end{bfseries} We use the DCC algorithm to learn each row of binary codes $B$ iteratively. Specifically, we suppose that $u^T,z^T$ and $q^T$ denote $l-th$ row of $W,B$ and $Q$ respectively. After excluding $u^T,z^T$ and $q^T$ from $W,B$ and $Q$, the resulting matrices are $\tilde{W}$,$\tilde{B}$ and $\tilde{Q}$ respectively. By using refashion strategy, we can obtain the closed solution (14). Please refer to \cite{38,39} for a more detailed introduction.
\begin{equation}
z=sgn(q-\tilde{B}^T\tilde{W}u)
\end{equation}
\hspace*{0.5cm}Each row of $B$ is updated iteratively by fixing the other $L-1$ rows. The integral optimization procedure of MFDH is summarized in Algorithm 1.\\
\hspace*{0.5cm}In the test phase, for a given new query $x$ (img or txt), we first compute multi-view kernelized features $x_1\in R^{d_1\times 1},x_2\in R^{d_2\times 1},x_3\in R^{d_3\times 1}$ according to Section 3.3, and then the hashing code $b$ of query $x$ is obtained as $b=sign(P_{img}X)$ for image or $b=sign(P_{txt}X)$ for text, where $X=[x_1^T,x_2^T,x_3^T]^T$.

\renewcommand{\algorithmicrequire}{\textbf{Input:}} 
\renewcommand{\algorithmicensure}{\textbf{Output:}}
\begin{algorithm}
\caption{Multi-view Features Discrete Hashing (MFDH)}
\begin{algorithmic}[1]
\REQUIRE image multi-view features ($Z^{img}$, $m^{img}$, $C^{img}$) and text multi-view features ($Z^{txt}$, $m^{txt}$, $C^{txt}$); $\alpha$, $\beta$, $\lambda$ and the length of hashing code ($L$); class label matrix $Y$.
\ENSURE Projection matrix $P_{img}$,$P_{txt}$ and $B$.\\
Calculate kernerlized features $\Psi$,$\Phi$ for image and text respectively.\\
Initialize $P_{img}$,$P_{txt}$ and $B\in R^{L\times n}$.
\REPEAT
\STATE  Optimize $P_{img}$ according to (10);\\
   \STATE  Calculate $P_{txt}$ according to (11);\\
    \STATE Calculate $W$ according to (13);
	\STATE Learn $B$ according to (14);
\UNTIL convergence
\end{algorithmic}
\end{algorithm}

\section{EXPERIMENTS AND ANALYSIS}
\subsection{Datasets}
\hspace*{0.5cm}We conduct experiments on four publicly available cross-modal datasets: WiKi \cite{35}, MMED \cite{paa4,paa5}, MIRFlickr \cite{36}, and NUS-WIDE \cite{37} respectively. These datasets, consisting of two modalities, i.e. image and text modalities, are widely applied in many previous works. The dimensions of multi-view features are set empirically. In our experiments, we employ the pre-trained word2vec model on Google corpus to transform every key word or tag in textual documents into a 300-dimensional word vector. In this way, each text can be represented by a set of vectors. As for the image features, the 128-dimensional dense SIFT features are extracted for each image of all datasets. Then, the first and second order textual features of each text are represented as a 300-dimensional mean vector and a $300\times 300$ covariance matrix respectively. Similarly, we can also get 128-dimensional first-order mean vector and $128\times128$ covariance matrix for each image. In addition, the histogram feature of each image from WiKi (MIRFLickr, NUS-WIDE and MMED) is extracted as 1000 (500, 500 and 500) dimensional feature vector by means of the BoVW model. Similarly, we also get the 50 (100, 100 and 100) dimensional feature histogram for each text on WiKi (MIRFLickr, NUS-WIDE and MMED). The statistics of all datasets are given in Table 1.\\
\begin{bfseries}Wiki\end{bfseries} \cite{35} contains 2,866 multimedia documents crawled from Wikipedia. Every document consists of a pair of an image and a text description, and every paired sample is classified as one of  10 categories. We take 2,866 pairs from the dataset to form the training set and the remaining data is used as a testing set\\
\begin{bfseries}MMED\end{bfseries} \cite{paa4,paa5} is collected from hundreds of news media sites and Flickr-like photo-sharing social media. It includes 412 real-world events covering various event categories, such as natural disaster, public security, protest, sport, election, festival, etc. We select samples belonging to natural disaster category from MMED to form a subset of 2476 image-text pairs. Each pair is classified into 12 categories, i.e. earthquake, floods, water crisis, storm, typhoon, wildfires, hurricane, tornado, heat wave, cyclone, tropical storm, snowstorm. 2000 pairs are randomly chosen from the subset as the training data and the remaining data as query data.\\
\begin{bfseries}MIRFlickr\end{bfseries} \cite{36} consists of 25,000 original images collected from the Flickr website. Each image and corresponding annotated tags constitute an image-text pair, and each pair is classified into some of 24 classes. We keep 10,729 paired samples whose text can generate at least five words after removing stopwords from the original dataset for our experiments. Then 6,500 paired samples are used as the training set, and the rest of the database as query set.\\
\begin{bfseries}NUS-WIDE \end{bfseries} \cite{37} comprises 269,648 images with 81 concepts collected from Flickr and each image is associated with annotated tags. The association is regarded as an image-text pair. We select 4,301 paired samples that belong to the top l0 most frequent concepts and have at least 20 tags. We randomly choose 2,534 pairs of the samples as the training data and 1,767 as the testing data.
\begin{table}
\centering
\caption{STATISTICS OF THREE STANDAR DATASETS}
\begin{tabular}{ccccc}
\hline\noalign{\smallskip}
DataSets & WiKi & MMED & MIRFlickr & NUS-WIDE \\
\noalign{\smallskip}\hline\noalign{\smallskip}
Data Set Size & 2866  & 2476 & 10729 & 4301\\
Training Set Size & 2173 & 2000 & 6500 & 2500 \\
Retrieval Set Size & 2173  & 2000 & 6500 & 2500\\
Query Set Size & 693 & 476 & 4229 & 1801 \\
Num. of Labels & 10 & 12 &  24 & 10 \\
\noalign{\smallskip}\hline
\end{tabular}
\end{table}

\begin{table}[h]
\centering
\caption{COMPARATIVE MAP RESULTS OF CROSS-MODAL RETRIEVAL TASKS ON WIKI. THE BEST PERFORMANCE IS MARKED BY BOLD.}
\begin{tabular}{|c|c|c|c|c|c|}
\hline
\multicolumn{2}{|c|}{\multirow{2}{*}{Methods/Datasets}}&
\multicolumn{4}{c|}{WiKi}\\
\cline{3 - 6}
\multicolumn{2}{|c|}{}
&16&32&64&128\\
\hline
\hline
\multirow{9}{*}{\makecell[tl]{Image\\Query\\Text\\(I2T)}}&CCA\cite{23}&0.1699&0.1519&0.1495&0.1472\\
\cline{2-6}
&SCM-Orth\cite{15}&0.1538&0.1402&0.1303&0.1289\\
\cline{2-6}
&SCM-Seq\cite{15}&0.2341&0.2410&0.2462&0.2566\\
\cline{2-6}
&GSePH\cite{35}&0.2367&0.2561&0.2811&0.2903\\
\cline{2-6}
&$SePH_{rnd}$\cite{40}&0.2468&0.2779&0.3062&0.3148\\
\cline{2-6}
&$SePH_{km}$\cite{40}&0.2497&0.2796&0.3025&0.3147\\
\cline{2-6}
&DCH\cite{39}&0.3465&0.3584&0.3737&0.3798\\
\cline{2-6}
&DCH+RBF\cite{39}&0.3309&0.3505&0.3758&0.3741\\
\cline{2-6}
&MFDH&\textbf{0.3548}&\textbf{0.3763}&\textbf{0.3878}&\textbf{0.3954}\\
\cline{1-6}

\multirow{9}{*}{\makecell[tl]{Text\\Query\\Image\\(T2I)}}&CCA\cite{23}&0.1587&0.1392&0.1272&0.1211\\
\cline{2-6}
&SCM-Orth\cite{15}&0.1540&0.1373&0.1258&0.1224\\
\cline{2-6}
&SCM-Seq\cite{15}&0.2257&0.2459&0.2485&0.2528\\
\cline{2-6}
&GSePH\cite{35}&0.6138&0.6307&0.6388&0.6522\\
\cline{2-6}
&$SePH_{rnd}$\cite{40}&0.6226&0.6446&0.6600&0.6710\\
\cline{2-6}
&$SePH_{km}$\cite{40}&0.6267&0.6474&0.6603&0.6734\\
\cline{2-6}
&DCH\cite{39}&0.7046&0.7159&0.7201&0.7222\\
\cline{2-6}
&DCH+RBF\cite{39}&0.6984&0.7162&0.7223&0.7229\\
\cline{2-6}
&MFDH&\textbf{0.8318}&\textbf{0.8458}&\textbf{0.8568}&\textbf{0.8666}\\
\cline{1-6}
\end{tabular}
\end{table}

%%%%%
\begin{table}[h]
\centering
\caption{COMPARATIVE MAP RESULTS OF CROSS-MODAL RETRIEVAL TASKS ON MMED. THE BEST PERFORMANCE IS MARKED BY BOLD.}
\begin{tabular}{|c|c|c|c|c|c|}
\hline
\multicolumn{2}{|c|}{\multirow{2}{*}{Methods/Datasets}}&
\multicolumn{4}{c|}{MMED}\\
\cline{3 - 6}
\multicolumn{2}{|c|}{}
&16&32&64&128\\
\hline
\hline
\multirow{9}{*}{\makecell[tl]{Image\\Query\\Text\\(I2T)}}&CCA\cite{23}&0.1691&0.1671&0.1642&0.1613\\
\cline{2-6}
&SCM-Orth\cite{15}&0.1988&0.1827&0.1738&0.1699\\
\cline{2-6}
&SCM-Seq\cite{15}&0.3077&0.3309&0.3336&0.3403\\
\cline{2-6}
&GSePH\cite{35}&0.3141&0.3156&0.3334&0.3264\\
\cline{2-6}
&$SePH_{rnd}$\cite{40}&0.3674&0.3732&0.3738&0.3912\\
\cline{2-6}
&$SePH_{km}$\cite{40}&0.3647&0.3720&0.3708&0.3916\\
\cline{2-6}
&DCH\cite{39}&0.3678&0.3753&0.3863&0.3860\\
\cline{2-6}
&DCH+RBF\cite{39}&0.3278&0.3980&0.4058&0.4076\\
\cline{2-6}
&MFDH&\textbf{0.4812}&\textbf{0.4827}&\textbf{0.4908}&\textbf{0.4934}\\
\cline{1-6}

\multirow{9}{*}{\makecell[tl]{Text\\Query\\Image\\(T2I)}}&CCA\cite{23}&0.1717&0.1692&0.1673&0.1650\\
\cline{2-6}
&SCM-Orth\cite{15}&0.1902&0.1744&0.1627&0.1595\\
\cline{2-6}
&SCM-Seq\cite{15}&0.2571&0.2859&0.2958&0.3004\\
\cline{2-6}
&GSePH\cite{35}&0.2297&0.2371&0.2461&0.2459\\
\cline{2-6}
&$SePH_{rnd}$\cite{40}&0.8317&0.8442&0.8643&0.8779\\
\cline{2-6}
&$SePH_{km}$\cite{40}&0.8236&0.8433&0.8643&0.8750\\
\cline{2-6}
&DCH\cite{39}&0.8602&0.8816&0.8912&0.8995\\
\cline{2-6}
&DCH+RBF\cite{39}&0.6901&0.9021&0.9035&0.9076\\
\cline{2-6}
&MFDH&\textbf{0.9166}&\textbf{0.9476}&\textbf{0.9490}&\textbf{0.9552}\\
\cline{1-6}
\end{tabular}
\end{table}

\begin{table}[h]
\centering
\caption{COMPARATIVE MAP RESULTS OF CROSS-MODAL RETRIEVAL TASKS ON MIRFlickr. THE BEST PERFORMANCE IS MARKED BY BOLD.}
\begin{tabular}{|c|c|c|c|c|c|}
\hline
\multicolumn{2}{|c|}{\multirow{2}{*}{Methods/Datasets}}&
\multicolumn{4}{c|}{MIRFlickr}\\
\cline{3 - 6}
\multicolumn{2}{|c|}{}
&16&32&64&128\\
\hline
\hline
\multirow{9}{*}{\makecell[tl]{Image\\Query\\Text\\(I2T)}}&CCA\cite{23}&0.5602&0.5580&0.5566&0.5555\\
\cline{2-6}
&SCM-Orth\cite{15}&0.5741&0.5670&0.5615&0.5602\\
\cline{2-6}
&SCM-Seq\cite{15}&0.6003&0.6112&0.6215&0.6275\\
\cline{2-6}
&GSePH\cite{35}&0.6346&0.6180&0.6381&0.6363\\
\cline{2-6}
&$SePH_{rnd}$\cite{40}&0.6442&0.6490&0.6537&0.6564\\
\cline{2-6}
&$SePH_{km}$\cite{40}&0.6499&0.6533&0.6578&0.6583\\
\cline{2-6}
&DCH\cite{39}&0.6439&0.6291&0.6137&0.6153\\
\cline{2-6}
&DCH+RBF\cite{39}&0.6360&0.6276&0.6066&0.6142\\
\cline{2-6}
&MFDH&\textbf{0.6836}&\textbf{0.6939}&\textbf{0.7066}&\textbf{0.7230}\\
\cline{1-6}

\multirow{9}{*}{\makecell[tl]{Text\\Query\\Image\\(T2I)}}&CCA\cite{23}&0.5578&0.5562&0.5553&0.5548\\
\cline{2-6}
&SCM-Orth\cite{15}&0.5721&0.5654&0.5603&0.5587\\
\cline{2-6}
&SCM-Seq\cite{15}&0.5855&0.5941&0.6022&0.6075\\
\cline{2-6}
&GSePH\cite{35}&0.7109&0.7226&0.7301&0.7303\\
\cline{2-6}
&$SePH_{rnd}$\cite{40}&0.7226&0.7384&0.7401&0.7491\\
\cline{2-6}
&$SePH_{km}$\cite{40}&0.7304&0.7411&0.7482&0.7476\\
\cline{2-6}
&DCH\cite{39}&0.7284&0.6985&0.6764&0.6737\\
\cline{2-6}
&DCH+RBF\cite{39}&0.7128&0.7052&0.6702&0.6694\\
\cline{2-6}
&MFDH&\textbf{0.7408}&\textbf{0.7506}&\textbf{0.7602}&\textbf{0.7797}\\
\cline{1-6}
\end{tabular}
\end{table}

\begin{table}[h]
\centering
\caption{COMPARATIVE MAP RESULTS OF CROSS-MODAL RETRIEVAL TASKS ON NUS-WIDE. THE BEST PERFORMANCE IS MARKED BY BOLD.}
\begin{tabular}{|c|c|c|c|c|c|}
\hline
\multicolumn{2}{|c|}{\multirow{2}{*}{Methods/Datasets}}&
\multicolumn{4}{c|}{NUS-WIDE}\\
\cline{3 - 6}
\multicolumn{2}{|c|}{}
&16&32&64&128\\
\hline
\hline
\multirow{9}{*}{\makecell[tl]{Image\\Query\\Text\\(I2T)}}&CCA\cite{23}&0.4936&0.4920&0.4910&0.4903\\
\cline{2-6}
&SCM-Orth\cite{15}&0.5305&0.5171&0.5065&0.5019\\
\cline{2-6}
&SCM-Seq\cite{15}&0.6319&0.6357&0.6415&0.6437\\
\cline{2-6}
&GSePH\cite{35}&0.6022&0.6009&0.6121&0.6064\\
\cline{2-6}
&$SePH_{rnd}$\cite{40}&0.6742&0.6710&0.6848&0.6785\\
\cline{2-6}
&$SePH_{km}$\cite{40}&\textbf{0.6784}&\textbf{0.6784}&0.6851&0.6799\\
\cline{2-6}
&DCH\cite{39}&0.6070&0.5995&0.5949&0.6239\\
\cline{2-6}
&DCH+RBF\cite{39}&0.6203&0.5997&0.5944&0.6267\\
\cline{2-6}
&MFDH&0.6460&0.6714&\textbf{0.7014}&\textbf{0.7121}\\
\cline{1-6}

\multirow{9}{*}{\makecell[tl]{Text\\Query\\Image\\(T2I)}}&CCA\cite{23}&0.4942&0.4936&0.4927&0.4921\\
\cline{2-6}
&SCM-Orth\cite{15}&0.5379&0.5191&0.5084&0.5029\\
\cline{2-6}
&SCM-Seq\cite{15}&0.6457&0.6534&0.6667&0.6437\\
\cline{2-6}
&GSePH\cite{35}&0.7115&0.7075&0.7344&0.7316\\
\cline{2-6}
&$SePH_{rnd}$\cite{40}&0.7875&0.7943&0.8037&0.8053\\
\cline{2-6}
&$SePH_{km}$\cite{40}&\textbf{0.7907}&0.8011&0.8114&0.8100\\
\cline{2-6}
&DCH\cite{39}&0.7098&0.6944&0.6825&0.7533\\
\cline{2-6}
&DCH+RBF\cite{39}&0.7421&0.7127&0.6987&0.7570\\
\cline{2-6}
&MFDH&0.7811&\textbf{0.8285}&\textbf{0.8653}&\textbf{0.8824}\\
\cline{1-6}
\end{tabular}
\end{table}

\begin{table}[h]
\centering
\caption{COMPARATIVE MAP RESULTS OF UNIMODAL RETRIEVAL TASKS ON WIKI.THE BEST PERFORMANCE IS MARKED BY BOLD.}
\begin{tabular}{|c|c|c|c|c|c|}
\hline
\multicolumn{2}{|c|}{\multirow{2}{*}{Methods/Datasets}}&
\multicolumn{4}{c|}{WiKi}\\
\cline{3 - 6}
\multicolumn{2}{|c|}{}
&16&32&64&128\\
\hline
\hline
\multirow{9}{*}{\makecell[tl]{Image\\Query\\Image\\(I2I)}}&CCA\cite{23}&0.1254&0.1207&0.1184&0.1176\\
\cline{2-6}
&SCM-Orth\cite{15}&0.1257&0.1200&0.1179&0.1165\\
\cline{2-6}
&SCM-Seq\cite{15}&0.1471&0.1506&0.1513&0.1519\\
\cline{2-6}
&SDH\cite{38}&0.1496&0.1568&0.1598&0.1632\\
\cline{2-6}
&$SePH_{rnd}$\cite{40}&0.1748&0.1893&0.2141&0.2175\\
\cline{2-6}
&$SePH_{km}$\cite{40}&0.1735&0.1915&0.2053&0.2132\\
\cline{2-6}
&DCH\cite{39}&0.1515&0.1584&0.1614&0.1639\\
\cline{2-6}
&DCH+RBF\cite{39}&0.1493&0.1548&0.1613&0.1638\\
\cline{2-6}
&MFDH&\textbf{0.2815}&\textbf{0.3133}&\textbf{0.3312}&\textbf{0.3333}\\
\cline{1-6}

\multirow{9}{*}{\makecell[tl]{Text\\Query\\Text\\(T2T)}}&CCA\cite{23}&0.4301&0.4333&0.4510&0.4523\\
\cline{2-6}
&SCM-Orth\cite{15}&0.3017&0.2595&0.2264&0.2186\\
\cline{2-6}
&SCM-Seq\cite{15}&0.5214&0.5400&0.5516&0.5476\\
\cline{2-6}
&SDH\cite{38}&0.5322&0.5597&0.5715&0.5786\\
\cline{2-6}
&$SePH_{rnd}$\cite{40}&0.5655&0.5841&0.5956&0.6058\\
\cline{2-6}
&$SePH_{km}$\cite{40}&0.5667&0.5830&0.5955&0.6049\\
\cline{2-6}
&DCH\cite{39}&0.5468&0.5617&0.5717&0.5788\\
\cline{2-6}
&DCH+RBF\cite{39}&0.5413&0.5651&0.5767&0.5789\\
\cline{2-6}
&MFDH&\textbf{0.8370}&\textbf{0.8464}&\textbf{0.8568}&\textbf{0.8646}\\
\cline{1-6}
\end{tabular}
\end{table}

%%%%
\begin{table}[h]
\centering
\caption{COMPARATIVE MAP RESULTS OF UNIMODAL RETRIEVAL TASKS ON MMED.THE BEST PERFORMANCE IS MARKED BY BOLD.}
\begin{tabular}{|c|c|c|c|c|c|}
\hline
\multicolumn{2}{|c|}{\multirow{2}{*}{Methods/Datasets}}&
\multicolumn{4}{c|}{MMED}\\
\cline{3 - 6}
\multicolumn{2}{|c|}{}
&16&32&64&128\\
\hline
\hline
\multirow{9}{*}{\makecell[tl]{Image\\Query\\Image\\(I2I)}}&CCA\cite{23}&0.1639&0.1628&0.1616&0.1600\\
\cline{2-6}
&SCM-Orth\cite{15}&0.1668&0.1611&0.1571&0.1564\\
\cline{2-6}
&SCM-Seq\cite{15}&0.1942&0.2000&0.2007&0.2020\\
\cline{2-6}
&SDH\cite{38}&0.1783&0.1793&0.1821&0.1804\\
\cline{2-6}
&$SePH_{rnd}$\cite{40}&0.2003&0.2060&0.2078&0.2140\\
\cline{2-6}
&$SePH_{km}$\cite{40}&0.1994&0.2043&0.2067&0.2132\\
\cline{2-6}
&DCH\cite{39}&0.1778&0.1795&0.1810&0.1814\\
\cline{2-6}
&DCH+RBF\cite{39}&0.1735&0.1797&0.1822&0.1842\\
\cline{2-6}
&MFDH&\textbf{0.4806}&\textbf{0.4833}&\textbf{0.4910}&\textbf{0.4939}\\
\cline{1-6}

\multirow{9}{*}{\makecell[tl]{Text\\Query\\Text\\(T2T)}}&CCA\cite{23}&0.1852&0.1834&0.1864&0.1923\\
\cline{2-6}
&SCM-Orth\cite{15}&0.3134&0.2686&0.2532&0.2535\\
\cline{2-6}
&SCM-Seq\cite{15}&0.5806&0.6632&0.7007&0.7110\\
\cline{2-6}
&SDH\cite{38}&0.7601&0.7931&0.8220&0.8303\\
\cline{2-6}
&$SePH_{rnd}$\cite{40}&0.8280&0.8412&0.8583&0.8718\\
\cline{2-6}
&$SePH_{km}$\cite{40}&0.8239&0.8400&0.8598&0.8667\\
\cline{2-6}
&DCH\cite{39}&0.7551&0.7953&0.8141&0.8273\\
\cline{2-6}
&DCH+RBF\cite{39}&0.8315&0.8775&0.8807&0.8851\\
\cline{2-6}
&MFDH&\textbf{0.9142}&\textbf{0.9459}&\textbf{0.9468}&\textbf{0.9537}\\
\cline{1-6}
\end{tabular}
\end{table}

\begin{table}[h]
\centering
\caption{COMPARATIVE MAP RESULTS OF UNIMODAL RETRIEVAL TASKS ON MIRFlickr.THE BEST PERFORMANCE IS MARKED BY BOLD.}
\begin{tabular}{|c|c|c|c|c|c|}
\hline
\multicolumn{2}{|c|}{\multirow{2}{*}{Methods/Datasets}}&
\multicolumn{4}{c|}{MIRFlickr}\\
\cline{3 - 6}
\multicolumn{2}{|c|}{}
&16&32&64&128\\
\hline
\hline
\multirow{9}{*}{\makecell[tl]{Image\\Query\\Image\\(I2I)}}&CCA\cite{23}&0.5589&0.5570&0.5592&0.5521\\
\cline{2-6}
&SCM-Orth\cite{15}&0.5594&0.5569&0.5553&0.5546\\
\cline{2-6}
&SCM-Seq\cite{15}&0.5736&0.5752&0.5784&0.5791\\
\cline{2-6}
&SDH\cite{38}&0.5774&0.5793&0.5800&0.5806\\
\cline{2-6}
&$SePH_{rnd}$\cite{40}&0.5970&0.5982&0.6013&0.6026\\
\cline{2-6}
&$SePH_{km}$\cite{40}&0.5994&0.6002&0.6036&0.6040\\
\cline{2-6}
&DCH\cite{39}&0.5782&0.5794&0.5801&0.5818\\
\cline{2-6}
&DCH+RBF\cite{39}&0.5785&0.5804&0.5793&0.5819\\
\cline{2-6}
&MFDH&\textbf{0.6304}&\textbf{0.6334}&\textbf{0.6499}&\textbf{0.6517}\\
\cline{1-6}

\multirow{9}{*}{\makecell[tl]{Text\\Query\\Text\\(T2T)}}&CCA\cite{23}&0.5596&0.5581&0.5574&0.5572\\
\cline{2-6}
&SCM-Orth\cite{15}&0.6240&0.5976&0.5776&0.5730\\
\cline{2-6}
&SCM-Seq\cite{15}&0.6446&0.6752&0.7025&0.7238\\
\cline{2-6}
&SDH\cite{38}&0.6682&0.6693&0.6733&0.6748\\
\cline{2-6}
&$SePH_{rnd}$\cite{40}&0.7134&0.7287&0.7312&0.7401\\
\cline{2-6}
&$SePH_{km}$\cite{40}&\textbf{0.7220}&\textbf{0.7335}&0.7404&0.7396\\
\cline{2-6}
&DCH\cite{39}&0.6745&0.6656&0.6634&0.6605\\
\cline{2-6}
&DCH+RBF\cite{39}&0.6773&0.6698&0.6510&0.6524\\
\cline{2-6}
&MFDH&0.6980&0.7301&\textbf{0.7487}&\textbf{0.7730}\\
\cline{1-6}
\end{tabular}
\end{table}

\begin{table}[h]
\centering
\caption{COMPARATIVE MAP RESULTS OF UNIMODAL RETRIEVAL TASKS ON NUS-WIDE.THE BEST PERFORMANCE IS MARKED BY BOLD.}
\begin{tabular}{|c|c|c|c|c|c|}
\hline
\multicolumn{2}{|c|}{\multirow{2}{*}{Methods/Datasets}}&
\multicolumn{4}{c|}{NUS-WIDE}\\
\cline{3 - 6}
\multicolumn{2}{|c|}{}
&16&32&64&128\\
\hline
\hline
\multirow{9}{*}{\makecell[tl]{Image\\Query\\Image\\(I2I)}}&CCA\cite{23}&0.4943&0.4919&0.4903&0.4892\\
\cline{2-6}
&SCM-Orth\cite{15}&0.5116&0.5023&0.4964&0.4937\\
\cline{2-6}
&SCM-Seq\cite{15}&0.5767&0.5845&0.5852&0.5884\\
\cline{2-6}
&SDH\cite{38}&0.5521&0.5498&0.5775&0.5842\\
\cline{2-6}
&$SePH_{rnd}$\cite{40}&0.6221&0.6232&0.6340&0.6285\\
\cline{2-6}
&$SePH_{km}$\cite{40}&0.6250&0.6236&0.6323&0.6290\\
\cline{2-6}
&DCH\cite{39}&0.5496&0.5590&0.5625&0.5741\\
\cline{2-6}
&DCH+RBF\cite{39}&0.5622&0.5480&0.5635&0.5752\\
\cline{2-6}
&MFDH&\textbf{0.6890}&\textbf{0.7031}&\textbf{0.7148}&\textbf{0.7230}\\
\cline{1-6}

\multirow{9}{*}{\makecell[tl]{Text\\Query\\Text\\(T2T)}}&CCA\cite{23}&0.4985&0.4970&0.4953&0.4952\\
\cline{2-6}
&SCM-Orth\cite{15}&0.5720&0.5508&0.5420&0.5381\\
\cline{2-6}
&SCM-Seq\cite{15}&0.7328&0.7510&0.7562&0.7607\\
\cline{2-6}
&SDH\cite{38}&0.6827&0.6768&0.7569&0.7651\\
\cline{2-6}
&$SePH_{rnd}$\cite{40}&0.7787&0.7843&0.7933&0.7943\\
\cline{2-6}
&$SePH_{km}$\cite{40}&0.7798&0.7927&0.8011&0.7962\\
\cline{2-6}
&DCH\cite{39}&0.6769&0.6860&0.6927&0.7235\\
\cline{2-6}
&DCH+RBF\cite{39}&0.7135&0.6836&0.6988&0.7303\\
\cline{2-6}
&MFDH&\textbf{0.7995}&\textbf{0.8343}&\textbf{0.8419}&\textbf{0.8546}\\
\cline{1-6}
\end{tabular}
\end{table}

\begin{table}
\centering
\caption{DIFFERENT COMBINATIONAL MODE OF KERNEL FUNCTION FOR MULTI-VIEW FEATURES,  THE SYMBOL '$\surd$' DENOTES CORRESPONDING KERNEL FUNCTION IS SELECTED.}
\begin{tabular}{lllll}
\hline\noalign{\smallskip}
\makecell[tl]{Serial Number \\of combination} &\makecell[tl]{ Kernel \\ function }& \makecell[tl]{ Histogram \\ feature }&\makecell[tl]{ 1st-order \\ feature }&\makecell[tl]{ 2nd-order \\ feature }\\
\hline
\multirow{2}{*}{\textcircled{1}}&RBF&$\surd$&$\surd$&$\surd$\\
\cline{2-5}
&Polynomial&&&\\
\hline
\multirow{2}{*}{\textcircled{2}}&RBF&&&\\
\cline{2-5}
&Polynomial&$\surd$&$\surd$&$\surd$\\
\hline
\multirow{2}{*}{\textcircled{3}}&RBF&$\surd$&$\surd$&\\
\cline{2-5}
&Polynomial&&&$\surd$\\
\hline
\multirow{2}{*}{\textcircled{4}}&RBF&$\surd$&&$\surd$\\
\cline{2-5}
&Polynomial&&$\surd$&\\
\hline
\multirow{2}{*}{\textcircled{5}}&RBF&&$\surd$&$\surd$\\
\cline{2-5}
&Polynomial&$\surd$&&\\
\hline
\multirow{2}{*}{\textcircled{6}}&RBF&$\surd$&&\\
\cline{2-5}
&Polynomial&&$\surd$&$\surd$\\
\hline
\multirow{2}{*}{\textcircled{7}}&RBF&&$\surd$&\\
\cline{2-5}
&Polynomial&$\surd$&&$\surd$\\
\hline
\multirow{2}{*}{\textcircled{8}}&RBF&&&$\surd$\\
\cline{2-5}
&Polynomial&$\surd$&$\surd$&\\
\hline
\end{tabular}
\end{table}

%%%%
\begin{table}[h]
\centering
\caption{The results of the ablation experiments on kernel stage.}
\begin{tabular}{|c|c|c|c}
\cline{1-3}
Dataset&Feature&Retrieval accuracy\\
\cline{1-3}
\multirow{4}{*}{\makecell[tl]{WiKi}}&$\square$&0.3239&\multirow{4}{*}{\makecell[tl]{$\square$ denotes that the model just uses \\ \quad the histogram feature and the \\ \quad 1st-order statistics feature}}\\
\cline{2-3}
&$\triangle$&0.6734&\\
\cline{2-3}
&$\otimes$&0.6855&\\
\cline{2-3}
&$\star$&\textbf{0.8568}&\\
\Xcline{1-3}{1.2pt}

\multirow{4}{*}{\makecell[tl]{MMED}}&$\square$&0.7032&\multirow{4}{*}{\makecell[tl]{$\triangle$ denotes that the model just uses \\ \quad the histogram feature and the \\ \quad 2nd-order statistics feature}}\\
\cline{2-3}
&$\triangle$&0.9011&\\
\cline{2-3}
&$\otimes$&0.9149&\\
\cline{2-3}
&$\star$&\textbf{0.9490}&\\
\Xcline{1-3}{1.2pt}

\multirow{4}{*}{\makecell[tl]{MIRFlickr}}&$\square$&0.6515&\multirow{4}{*}{\makecell[tl]{\quad $\otimes$ denotes that the model just uses\\ \quad \quad  the 1st-order statistics feature and\\ \quad \quad the 2nd-order statistics feature}}\\
\cline{2-3}
&$\triangle$&0.7099&\\
\cline{2-3}
&$\otimes$&0.6908&\\
\cline{2-3}
&$\star$&\textbf{0.7602}&\\
\Xcline{1-3}{1.2pt}

\multirow{4}{*}{\makecell[tl]{NUS-WIDE}}&$\square$&0.6952&\multirow{4}{*}{\makecell[tl]{$\star$ denotes that the model uses the \\ \quad histogram feature, 1st-order \\ \quad statistics feature and  2nd-order \\ \quad statistics feature}}\\
\cline{2-3}
&$\triangle$&0.8318&\\
\cline{2-3}
&$\otimes$&0.8403&\\
\cline{2-3}
&$\star$&\textbf{0.8653}&\\
\cline{1-3}

\end{tabular}
\end{table}

%%%%
\begin{table}[h]
\centering
\caption{The results of the ablation experiments on the terms in Equation 7.}
\begin{tabular}{|c|c|c|c|c|c|}
\hline
Dataset&&I2T&T2I&I2I&T2T\\
\hline
\multirow{4}{*}{\makecell[tl]{WiKi}}&MFDH($\alpha=0$)&0.1458&0.1129&0.1123&\textbf{0.8244}\\
\cline{2 - 6}
&MFDH($\beta=0$)&0.1115&0.1522&\textbf{0.3191}&0.1350\\
\cline{2 - 6}
&MFDH($\alpha,\beta$)&0.1126&0.1203&0.1127&0.1202\\
\cline{2 - 6}
&MFDH&0.3878&0.8568&0.3312&0.8568\\
\hline
\hline

\multirow{4}{*}{\makecell[tl]{MMED}}&MFDH($\alpha=0$)&0.2036&0.1520&0.1519&\textbf{0.9393}\\
\cline{2 - 6}
&MFDH($\beta=0$)&0.1576&0.1619&\textbf{0.4857}&0.1541\\
\cline{2 - 6}
&MFDH($\alpha,\beta$)&0.1531&0.1564&0.1530&0.1567\\
\cline{2 - 6}
&MFDH&0.4954&0.9490&0.4960&0.9077\\
\hline
\hline

\multirow{4}{*}{\makecell[tl]{MIRFlickr}}&MFDH($\alpha=0$)&0.5778&0.5450&0.5537&\textbf{0.7543}\\
\cline{2 - 6}
&MFDH($\beta=0$)&0.5504&0.5525&\textbf{0.6563}&0.5584\\
\cline{2 - 6}
&MFDH($\alpha,\beta$)&0.5516&0.5524&0.5585&0.5515\\
\cline{2 - 6}
&MFDH&0.7066&0.7602&0.6499&0.7487\\
\hline
\hline

\multirow{4}{*}{\makecell[tl]{NUS-WIDE}}&MFDH($\alpha=0$)&0.5371&0.4461&0.4906&\textbf{0.8470}\\
\cline{2 - 6}
&MFDH($\beta=0$)&0.4985&0.5140&\textbf{0.7230}&0.5713\\
\cline{2 - 6}
&MFDH($\alpha,\beta$)&0.4860&0.4873&0.4855&0.4892\\
\cline{2 - 6}
&MFDH&0.7014&0.8653&0.7148&0.8419\\
\hline
\end{tabular}
\end{table}

%%%%
\begin{table}[h]
\centering
\caption{The MAP comparison between Multi-view features and CNN feature.}
\begin{tabular}{|c|c|c|}
\hline
Dataset\verb|\|Feature Type&CNN Visual Feature&Multi-view features\\
\hline
WiKi&0.2658&\textbf{0.3312}\\
\cline{1 - 3}
MMED&\textbf{0.5323}&0.4960\\
\cline{1 - 3}
MIRFlickr&\textbf{0.7126}&0.6499\\
\cline{1 - 3}
NUS-WIDE&\textbf{0.7233}&0.7148\\
\cline{1 - 3}
Average accuracy&\textbf{0.5585}&0.5480\\
\cline{1 - 3}
\end{tabular}
\end{table}

\subsection{Experimental Setting}
\hspace*{0.5cm}MFDH proposed in this paper is compared with eight state-of-the-art cross-modal hashing methods, including CCA \cite{23}, SCM-Orth \cite{15}, SCM-Seq \cite{15}, GSePH \cite{35}, DCH \cite{39}, DCH+RBF \cite{39}, $SePH_{rnd}$ \cite{40,41} and $SePH_{km}$ \cite{40,41} respectively. The source codes of the compared methods except for GSePH are kindly provided by the original authors. Since the source code of GSePH is not publicly available, we implemented it by ourselves. Our experiments are executed on MATLAB 2016b on a Windows 10 (64-Bit) platform based desktop machine with 12 GB memory and 4-core 3.6GHz CPU. The CPU architectures are Intel(R) CORE(TM) i7-7700. $\lambda$ is the regularization coefficient in the objective function (7) to avoid overfitting. In experiments, $\lambda$ is set to be 0.01 empirically. Some amateur experiments are carried out by ourselves to determine the numerical values of $a$, $s$ and $\sigma$ in formula (3) and (4). In larger candidate range, we found that the case where $\sigma$, $a$ and $s$ are set into 1, 1 and 5 respectively is preferable. The baselines are designed based on a single-view feature. For WiKi, MIRFlickr, and NUS-WIDE, the baselines use uniformly the hand-crafted feature reported in the reference \cite{39} as the input of their models. It should be noted that the number of the training samples is different from one reported in the reference \cite{39} to ensure that the raw samples of the training set and testing set are consistent with MFDH proposed in this paper. On the MMED dataset, the first-order feature is extracted as the original feature of the baselines. The average values obtained by performing 10 runs for all the methods are recorded in this paper. \\
\subsubsection{Evaluation Metric}
\hspace*{0.5cm}The Mean Average Precision (MAP) is used as a criterion of retrieval performance. The Average Precision (AP) for a query $q$ is defined as E.q.(15) \\
\begin{equation}
AP(q) =\frac{1}{l_q}\sum_{m=1}^RP_q(m)\delta_q(m)
\end{equation}
where $l_q$ denotes the correct statistics of top $R$ retrieval results; $P_q(m)$ is the accuracy of top m retrieval results; and if the result of position $m$ is right,$\delta_q(m)$ equals one and is set to zero otherwise. The average value of AP of all queries is the final MAP. A larger MAP value indicates better performance.\\
\hspace*{0.5cm}Another widely-used criteria, i.e. Hash lookup, is adopted in our experiments. A lookup table is constructed to return the points that fall within a small Hamming radius $r$ of the query point. In our experiments, we plot the precision and recall curve according to the hash lookup protocol and varying the Hamming radius $r$. All query points are used to evaluate the performance. 
\begin{figure*}[htbp]
\subfigure[I2T]{
\includegraphics[width=.5\textwidth]{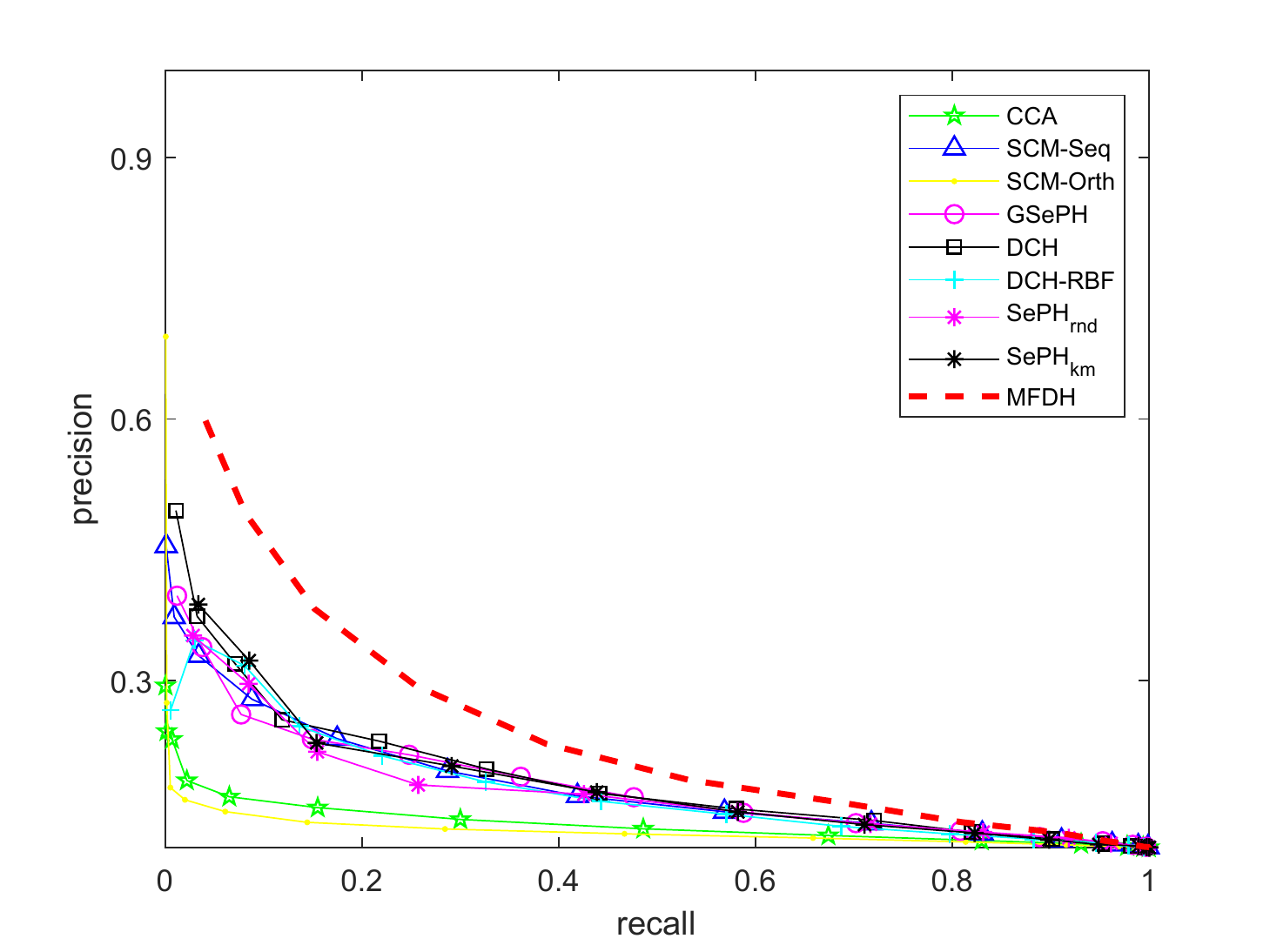}}
\subfigure[T2I]{
\includegraphics[width=.5\textwidth]{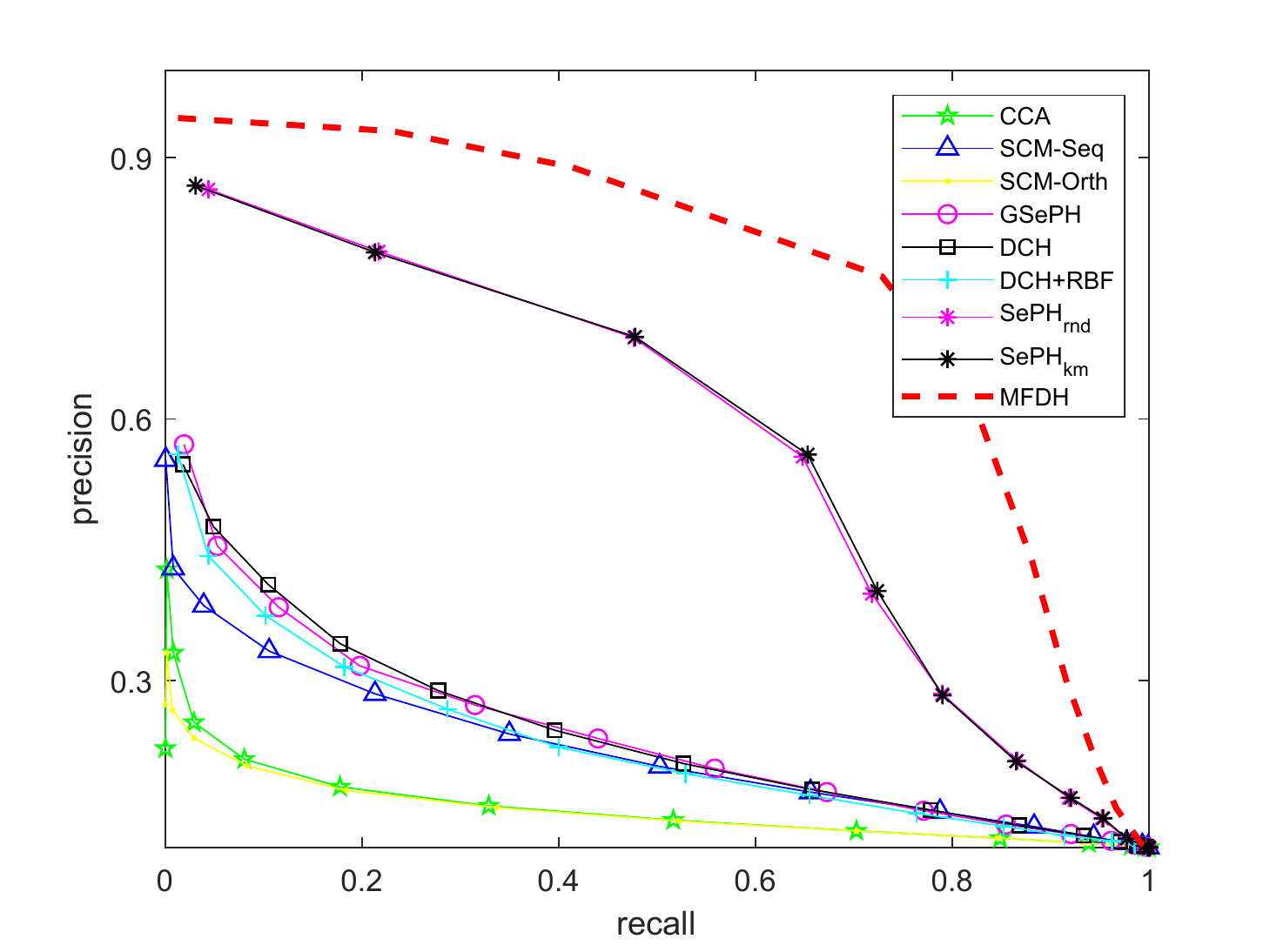}}
\caption{The precision-recall curves of all methods with 128-bit on WiKi.(a) I2T, (b)T2I}
\end{figure*}

\begin{figure*}[htbp]
\subfigure[I2T]{
\includegraphics[width=.5\textwidth]{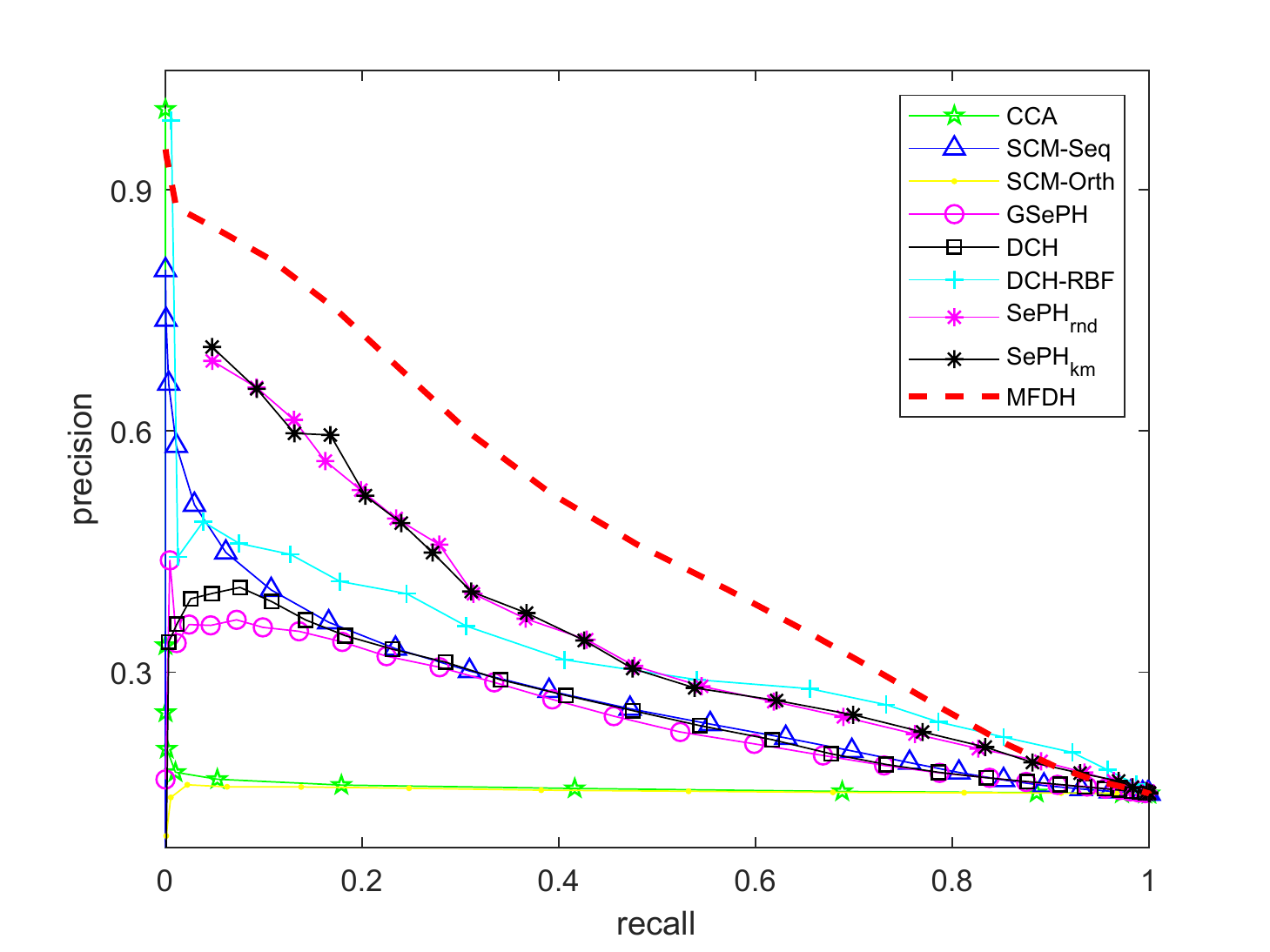}}
\subfigure[T2I]{
\includegraphics[width=.5\textwidth]{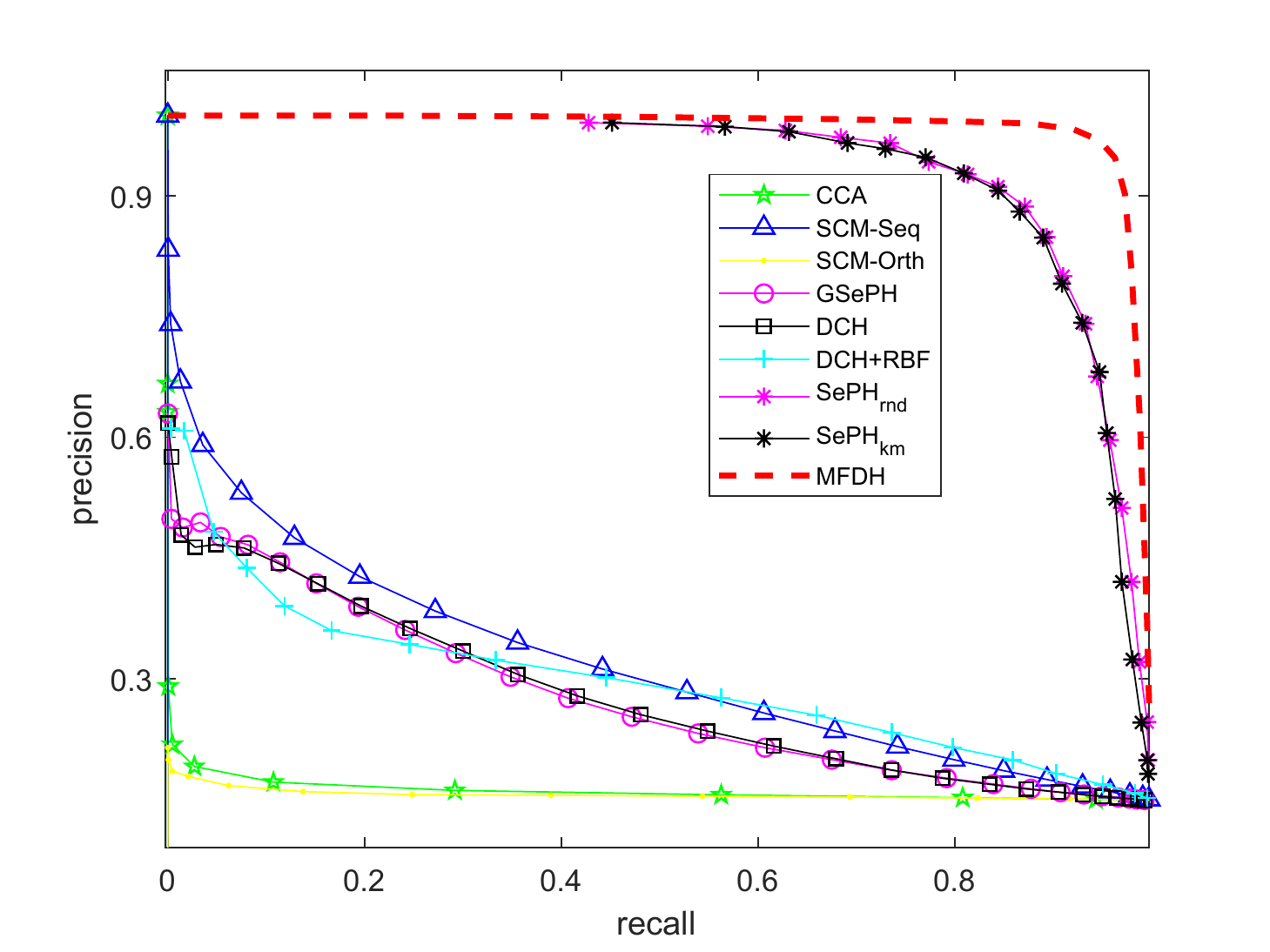}}
\caption{The precision-recall curves of all methods with 128-bit on MMED.(a) I2T, (b)T2I}
\end{figure*}

\begin{figure*}[htbp]
\subfigure[I2T]{
\includegraphics[width=.5\textwidth]{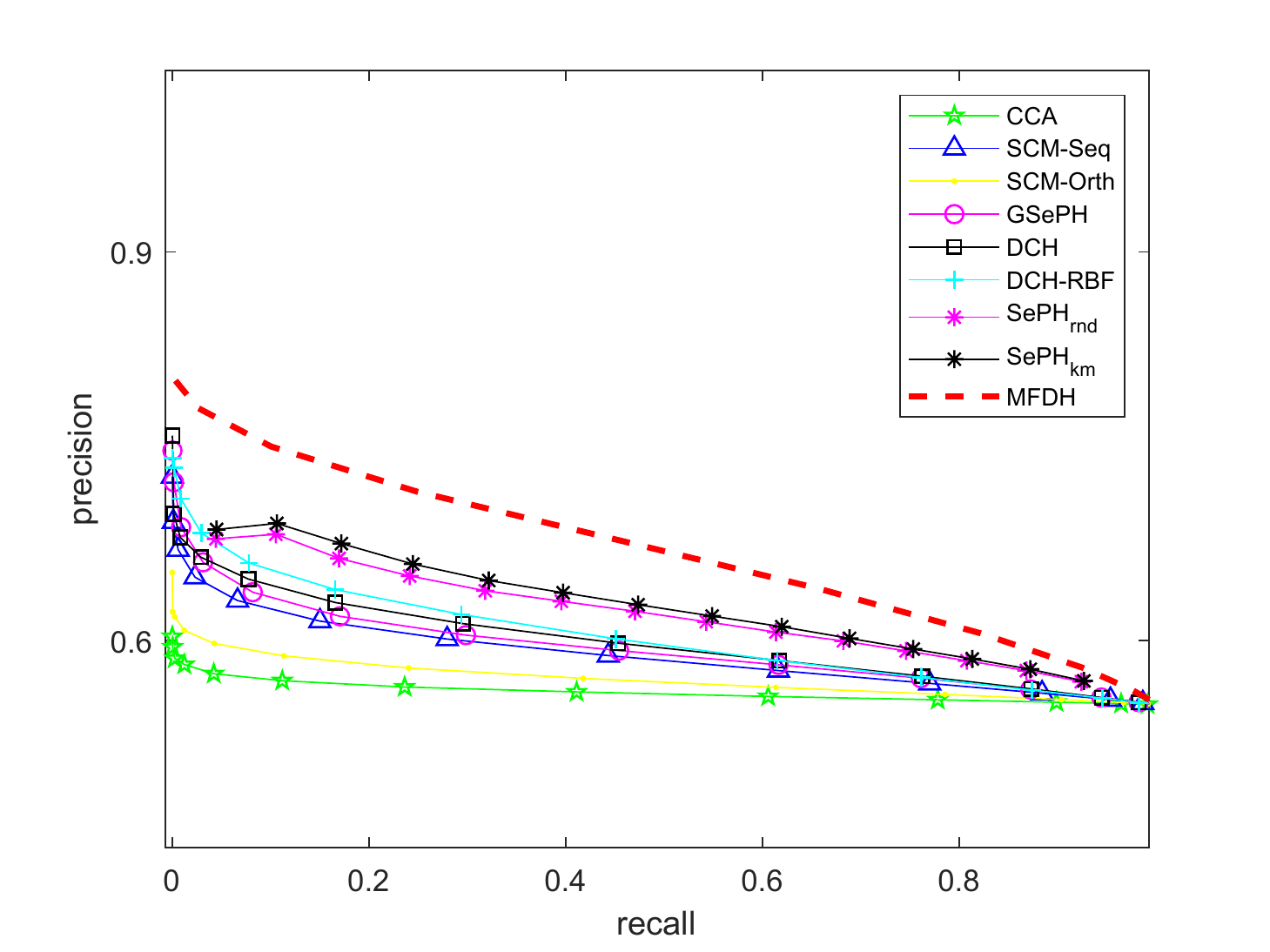}}
\subfigure[T2I]{
\includegraphics[width=.5\textwidth]{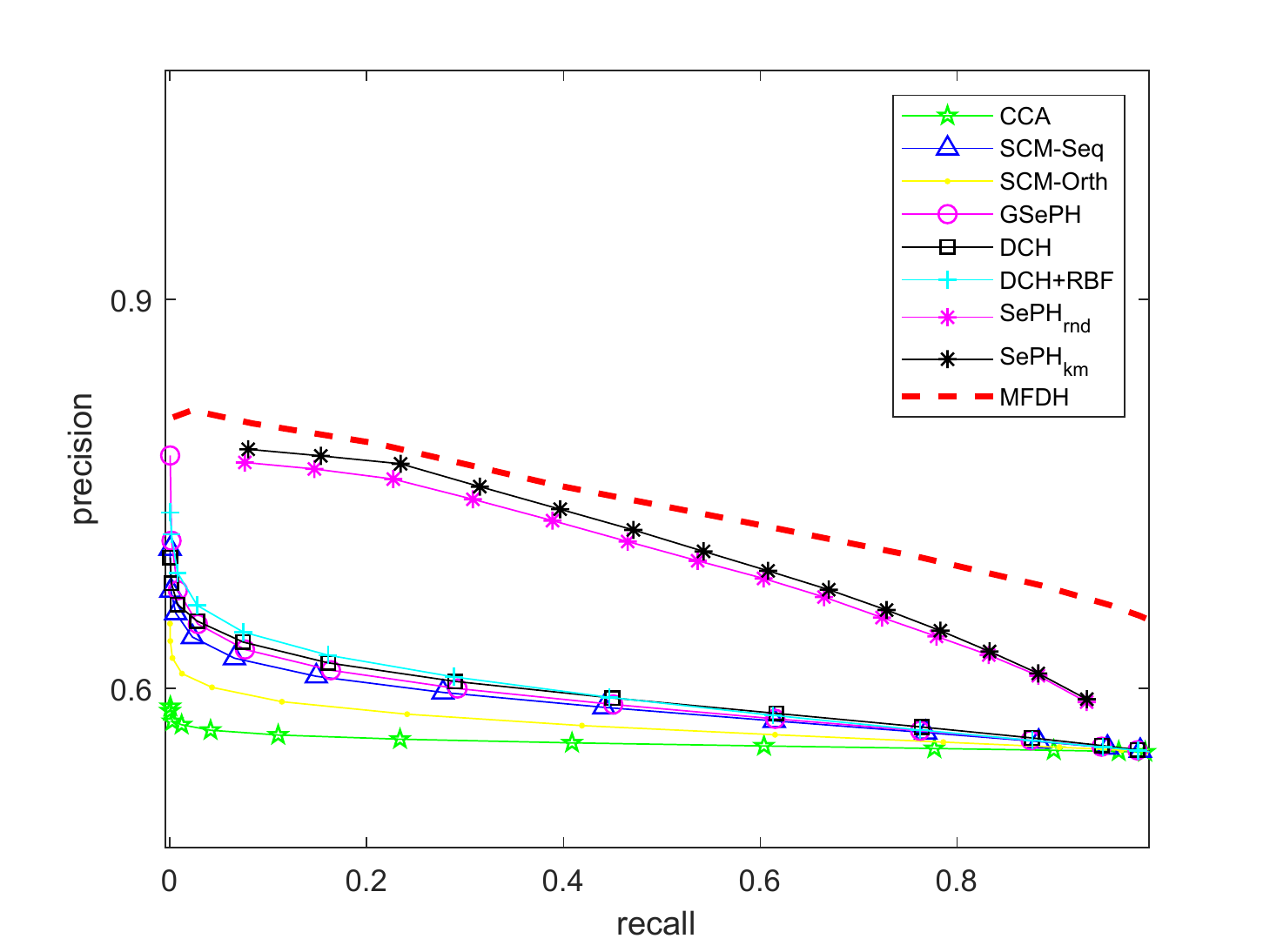}}
\caption{The precision-recall curves of all methods with 128-bit on MIRFlickr.(a) I2T, (b)T2I}
\end{figure*}

\begin{figure*}[htbp]
\subfigure[I2T]{
\includegraphics[width=.5\textwidth]{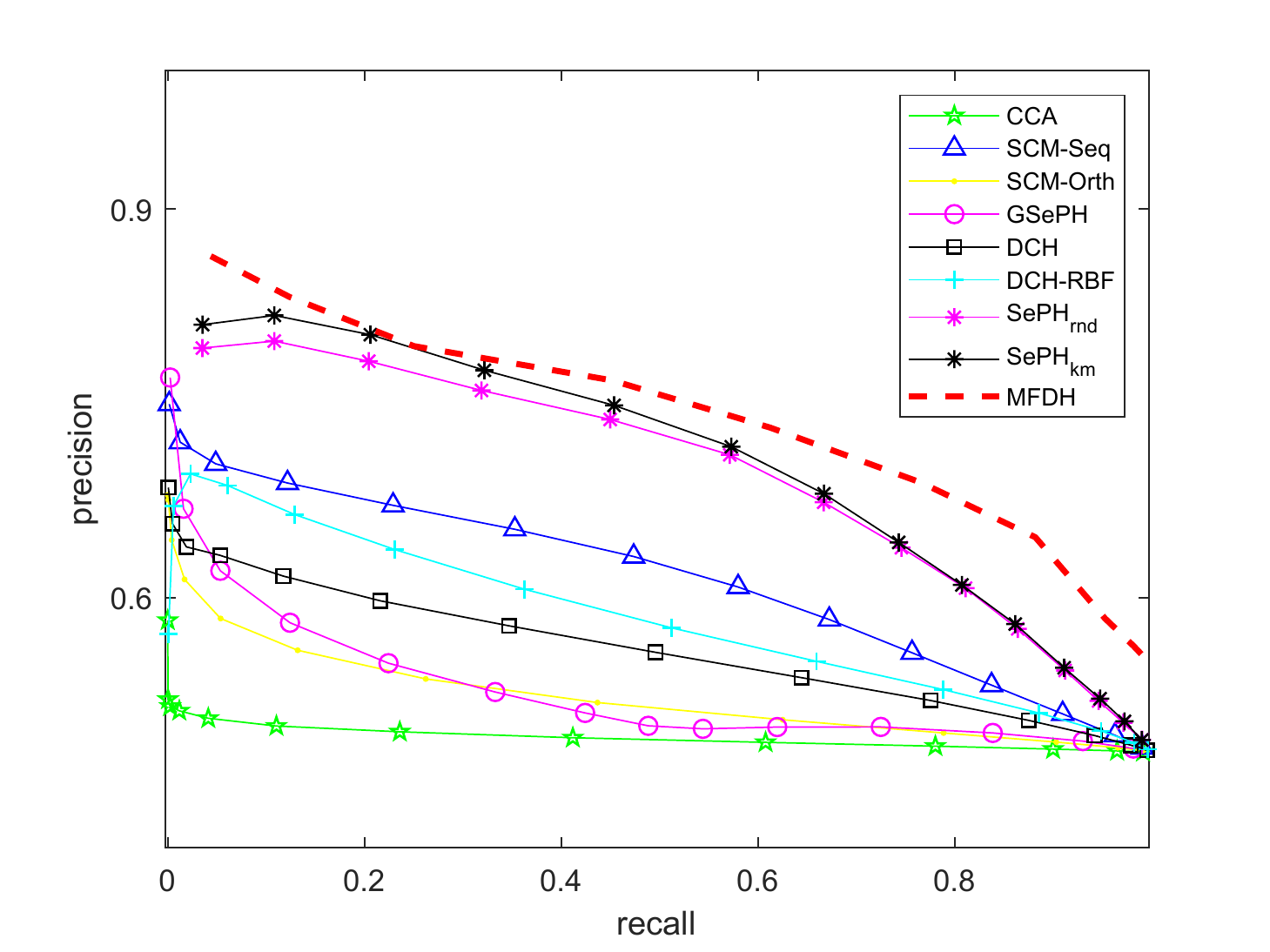}}
\subfigure[T2I]{
\includegraphics[width=.5\textwidth]{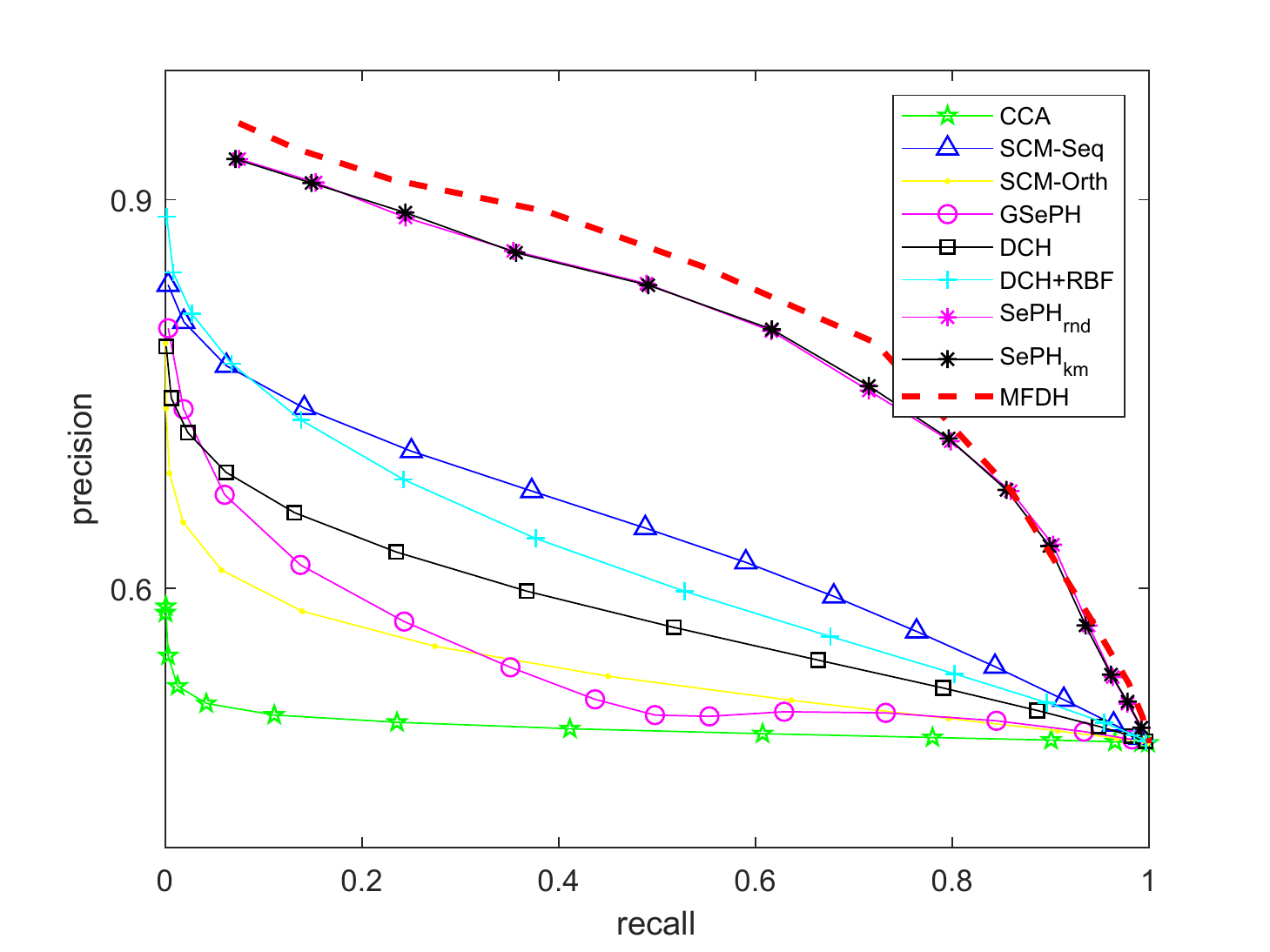}}
\caption{The precision-recall curves of all methods with 128-bit on NUS-WIDE.(a) I2T, (b)T2I}
\end{figure*}

\begin{figure*}[htbp]
%\centering
\subfigure[WiKi]{
\includegraphics[width=.5\textwidth]{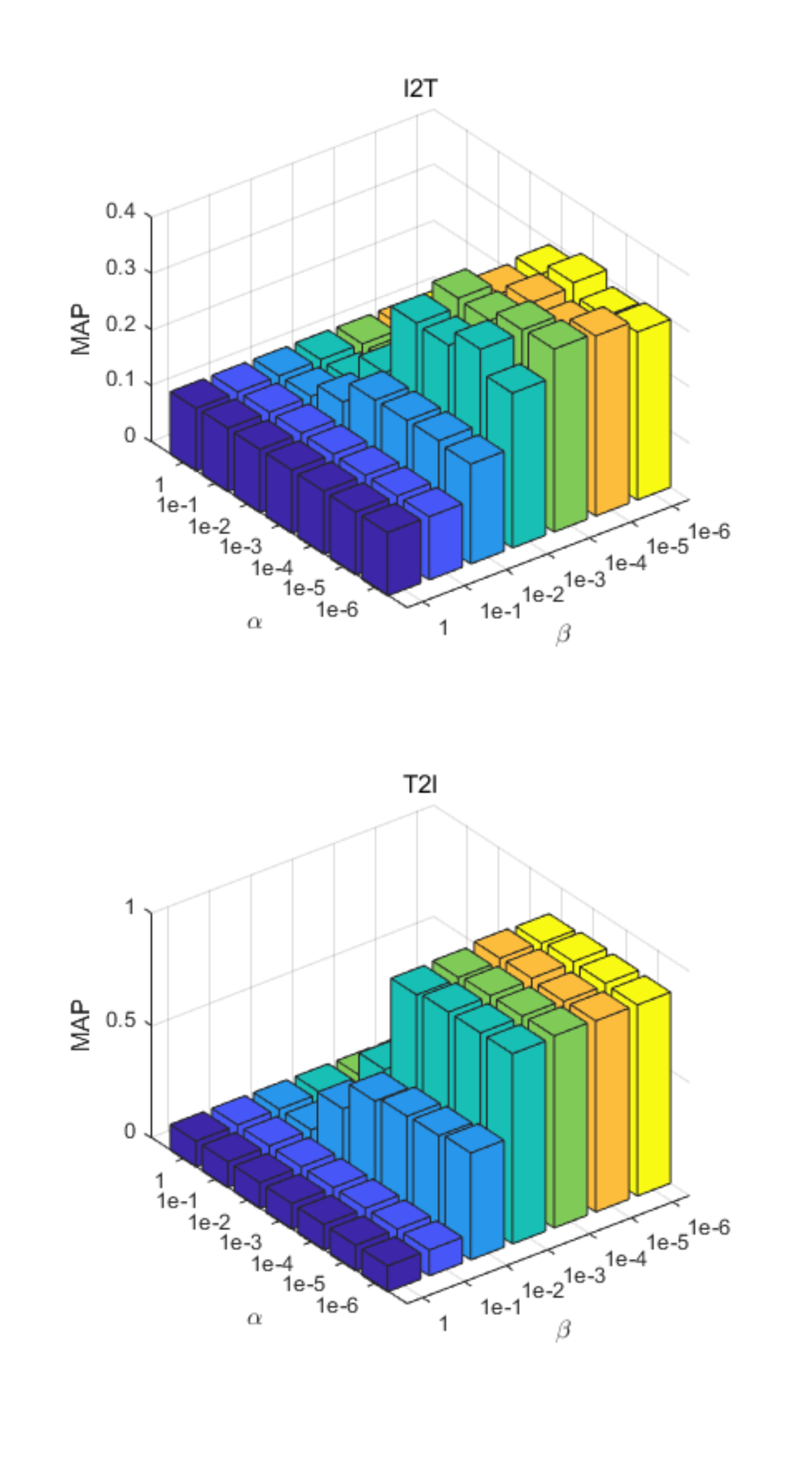}}
\subfigure[MMED]{
\includegraphics[width=.5\textwidth]{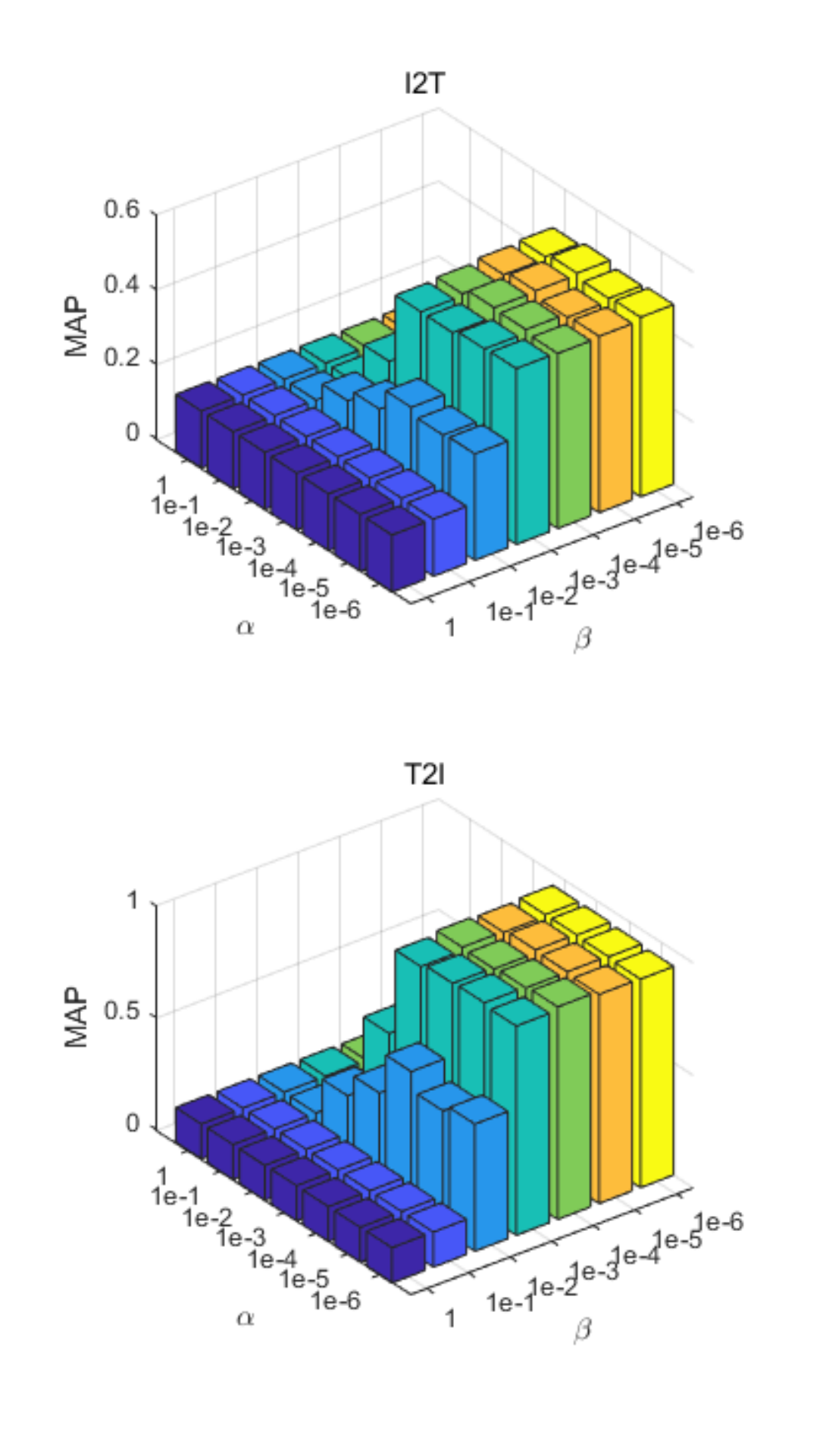}}
\subfigure[MIRFlickr]{
\includegraphics[width=.5\textwidth]{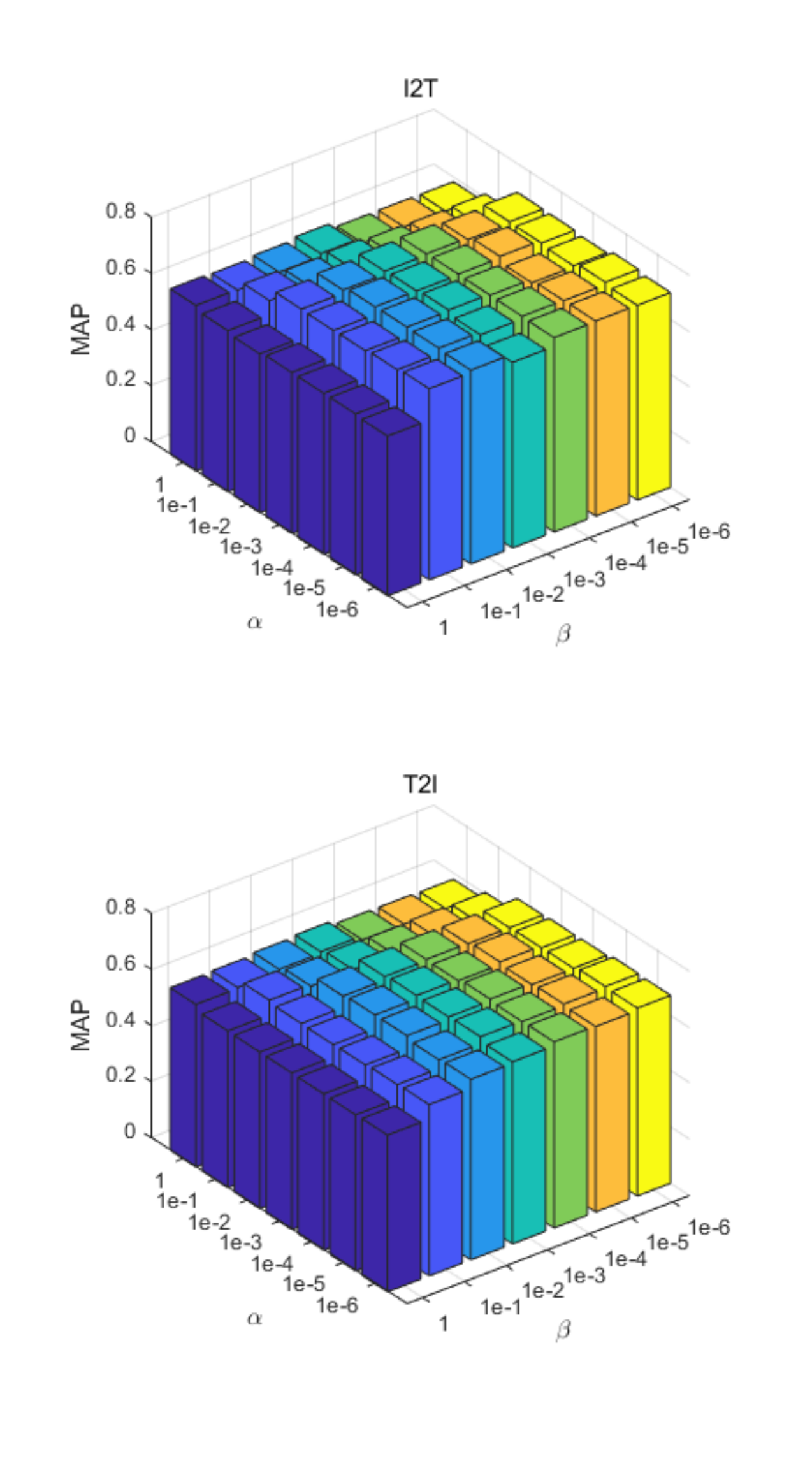}}
\subfigure[NUS-WIDE]{
\includegraphics[width=.5\textwidth]{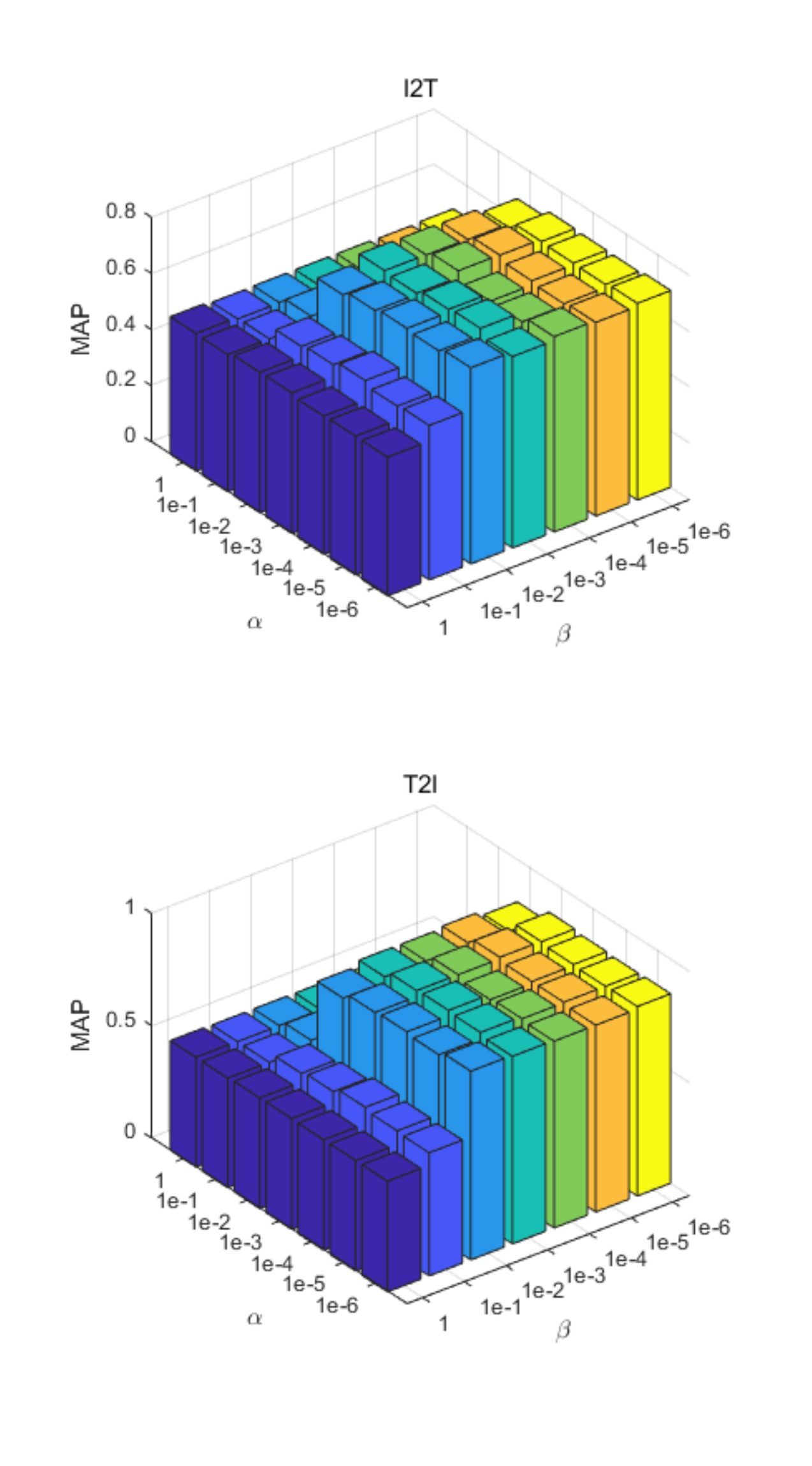}}
\caption{The performance variation of the MFDH with respect to $\alpha$ and $\beta$ using 128 bits on the WiKi (a), MMED (b),  MIRFlickr (c), and NUS-WIDE (d). }
\end{figure*}

\subsection{Retrieval results on Single-Label datasets} 
\hspace*{0.5cm}We carry out cross-modal retrieval including two typical retrieval tasks. i.e. Image querying Text and Text querying Image which are often abbreviated to I2T and T2I respectively in this paper. In particular, the meaning of I2T is as follows: Given an image search for texts of similar content from the dataset. By analogy, in T2I given a text as query to search for images of corresponding content in the dataset. In this section, we report the experimental results on the two single-label datasets (WiKi and MMED) where each sample associates a single label. As shown in Table 2 and Table 3, we can see that the proposed MFDH outperforms all compared methods on both tasks when the code length ranges from 16 bits to 128 bits. Specifically, MFDH achieves average improvements of 8\% and 7.2\% over the best baselines on WiKi and MMED respectively. In our experiments, we calculate the precision and recall according to the hash lookup protocol varying the Hamming radius. Fig.3 and Fig.4 plot the precision-recall curves with 128 bits hash codes on WiKi and MMED respectively. As illustrated in Fig. 3 and Fig.4, the scheme of MFDH consistently performs better than compared methods.\\ 
\hspace*{0.5cm}We extend our model to unimodal retrieval which involves Image query Image task and Text query Text task. Likewise, we use the acronyms I2I and T2T to represent the two tasks in this paper. As shown in Table 6 and Table 7, the proposed MFDH consistently achieves the best performance on all datasets comparing with the methods listed in Section 5.2. These results indicate that MFDH can generate effective hash codes.\\
\subsection{Retrieval results on Multi-Label datasets}
\hspace*{0.5cm}We testify our method on the MIRFlickr and NUS-WIDE to ensure that our model is flexible with multi-label semantic supervision. We perform similar experiments to Section 5.3 but on multi-label datasets whose multimodal data is associated with multiple labels. The comparative MAP results on the two datasets are shown in Table 4 and Table 5. We can observe that our MFDH is much better than the compared algorithms in term of MAP. Fig.5 and Fig.6 present the precision and recall curves on the MIRFlickr and NUS-WIDE respectively when fixing the length of hash codes into 128-bit. The shape of the precision and recall curves with other hash code length is very similar to one with 128 bit. As seen in Fig.5 and Fig.6, our MFDH  is superior to the compared hashing methods. For unimodal retrieval tasks, we conduct some comparative experiments under the multi-label scenario to evaluate the performance of our approach. The experimental results on MIRFlickr and NUS-WIDE are summarized in Table 8 and Table 9 respectively. We can easily observe that our approach outperforms other methods by a large performance margin. The promising results indicate our method with multi-view features is very effective under the multi-label data scenario.

\subsection{Parameter Sensitivity}
\hspace*{0.5cm}The two parameters  $\alpha$, $\beta$ in Equattion 7 control the quantization error of two modalities respectively. We vary the values of  $\alpha$, $\beta$ in the candidate range of \{$10^{-6}$, $10^{-5}$, $10^{-4}$, $10^{-3}$, $10^{-2}$, $10^{-1}$, 1\}. Fig.7 plots the performance of MFDH I2T and T2I as a function of  $\alpha$ and $\beta$ for a hashing code of 128 bits. The value of $\alpha$ and $\beta$ can be tuned manually. In Fig.7, we can see that the MFDH can achieve stable MAP result in a large range on WiKi, MMED, MIRFlickr and NUS-WIDE, which shows our model has weak dependence on the selectable parameters.
% Note that the point , $\alpha=\beta=0$ corresponds to MFDH only employing supervised label information to learn the discriminative hashing codes and ignoring the information provided by the image and text modalities. The solution just takes the information of one model into account when $\alpha$=0 or $\beta$=0. Thus, MFDH can not obtain the best performance in the above two cases.
\begin{figure*}
  \includegraphics[width=\textwidth]{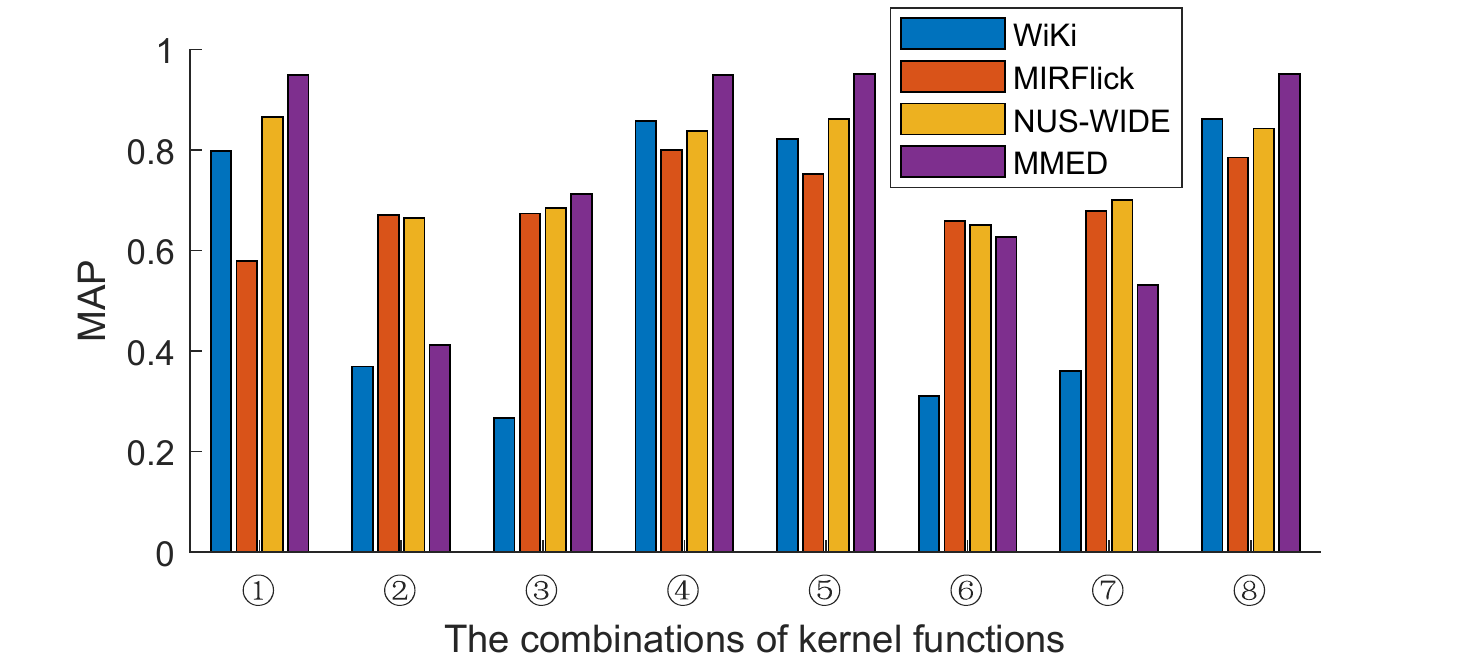}
\caption{The performance variation of our model using different combinations of kernel functions on WiKi, MIRFlickr, NUS-WIDE and MMED. The abscissa depicts different combinations of kernel functions (see Table 10).}
\end{figure*}

\begin{figure*}[htbp]
\subfigure[WiKi]{
\includegraphics[width=.5\textwidth]{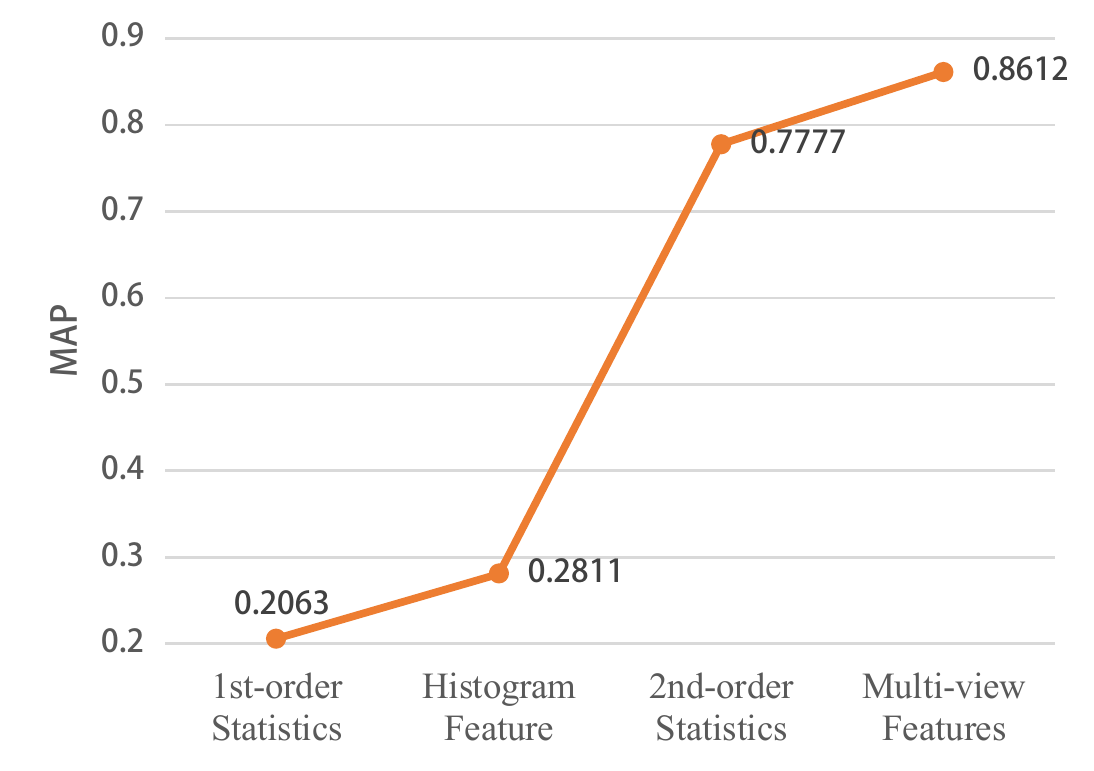}}
\subfigure[MMED]{
%\centering
\includegraphics[width=.5\textwidth]{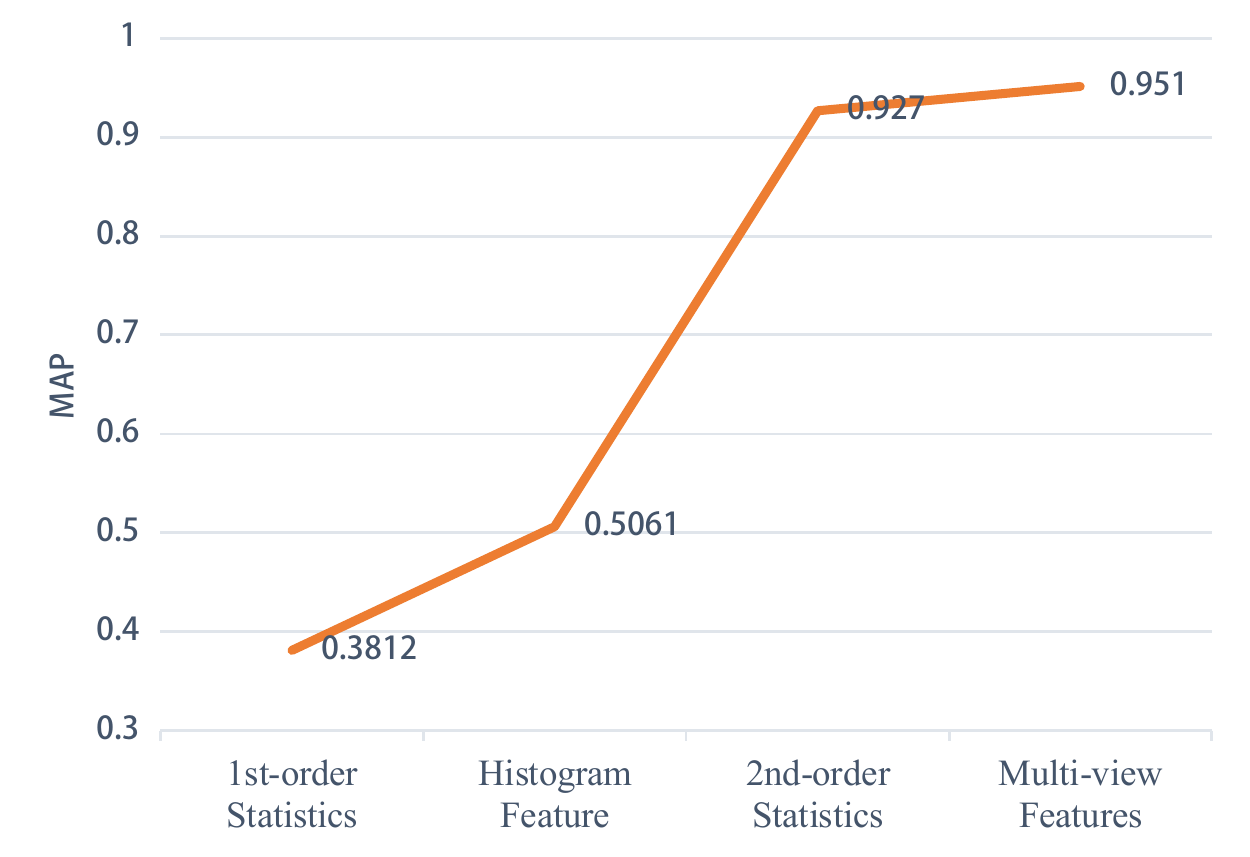}}
\subfigure[MIRFlickr]{
\includegraphics[width=.5\textwidth]{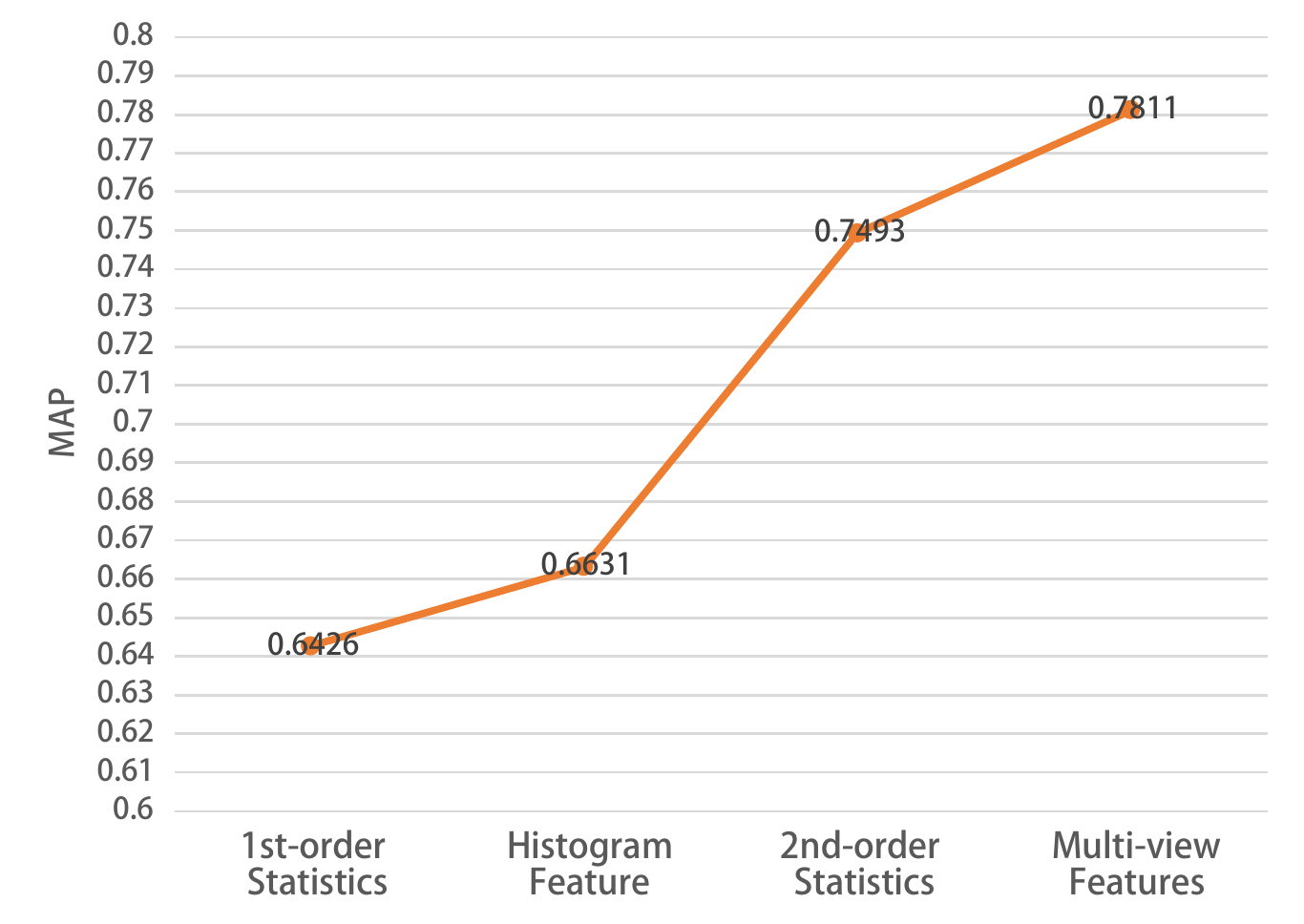}}
\subfigure[NUS-WIDE]{
%\centering
\includegraphics[width=.5\textwidth]{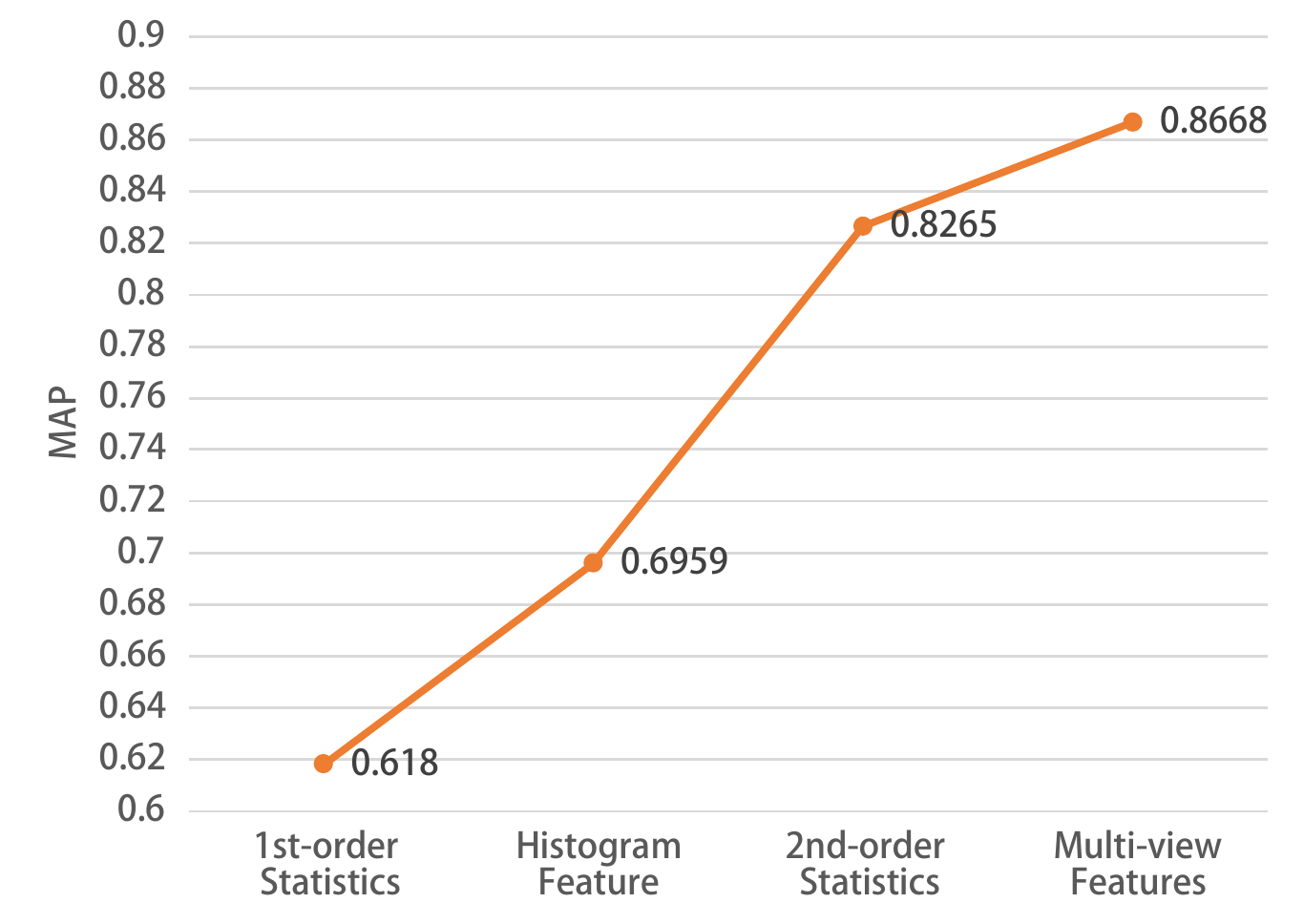}}
\caption{  The performance comparison of our model when using a single view feature and multi-view features respectively on WiKi  (a), MMED (b), MIRFlickr (c), and NUS-WIDE (d). }
\end{figure*}
\subsection{Discussion}
\subsubsection{Comparison with the baselines}
\hspace*{0.5cm}Some comparative experiments are performed to validate the effectiveness of the proposed method in this paper for two data scenarios presented in Section 5.3 and Section 5.4 respectively. We compare our method with the state-of-the-art hashing algorithms on four retrieval tasks. CCA \cite{23} is an unsupervised method in which the correlation between two modalities is maximized. SCM-Orth \cite{15} and SCM-Seq \cite{15} are two supervised methods which construct the semantic similarity using the label vectors. The common goal of the two methods is to maximize the semantic correlation. SCM-Orth \cite{15} imposes orthogonality constraint in the process of learning the hashing function while SCM-Seq \cite{15} utilizes a sequential strategy without orthogonality constraint. SePH \cite{40}, i.e. $SePH_{rnd}$ and $SePH_{km}$, transforms the semantic affinities into a probability distribution which is approximated in Hamming space. GSePH \cite{r8} first learns the optimal hash codes by preserving the semantic similarity between the data points and then learns the hash functions for two modalities by using kernel logistic regression. SDH \cite{38} and DCH \cite{39} are very efficient methods which use the ground truth labels to learn binary code. DCH introduces the DCC algorithm to optimize the problem with discrete constraint. DCH+RBF is an RBF kernel-based method for a nonlinear embedding. The main difference between DCH and DCH+RBF is that DCH+RBF modifies the linear embedding function of DCH to nonlinear to learn a common subspace. We can observe that DCH+RBF can not obtain very competing results compared with DCH. The possible reason why the performance of DCH+RBF is limited is that the a large amount of information has been lost in process of obtaining the original feature of samples, and the available discriminative information is insufficient. Considering the limited representation ability of single view feature, we design MFDH in the kernel stage fusing multiple views to solve this problem. Our framework employs multi-view features to represent images and texts for enriching the feature information and the complementary information among multiple views is exploited to better learn the discriminative compact hash code. The experimental results on four public datasets show our approach based on multi-view features can dramatically improve the retrieval accuracy.\\

\subsubsection{Comparison with CNN Visual feature}
\hspace*{0.5cm}Deep learning can naturally encode effective feature representation by nonlinear deep neural networks, which has shown state-of-the-art performance in retrieval applications. In experiments, the CNN visual feature is extracted from the model \cite{paa6} which is pre-trained on ImageNet. We use the CNN visual feature to replace the Multi-view features. Table 13 displays the retrieval results when using CNN Visual Feature and Multi-view features respectively on WiKi, MIRFlickr, NUS-WIDE, and MMED.  We can observe that the average performance of our model based on Multi-view features is very close to one based on CNN visual feature. On the WiKi, the performance when using Multi-view features is higher than the CNN visual feature by 6.5\%.  The experimental results shown in Table 13 demonstrate that multi-view features are very effective to represent multimodal data.

\subsubsection{Discussion on Different Combination of Kernel Functions}
\hspace*{0.5cm}The MFDH proposed in the paper selects two widely-used kernel functions (see E.q. (3) and E.q. (4)) to calculate kernel matrices. For the same view features of the two modalities (i.e. image and text), we adopt the same kernel function. As seen in Table 10, two kinds of kernel functions and three views construct eight combination modes. The MAP performance for different combinations of kernel functions (see Table 10) is shown in Fig.8.  In Fig.8, we can see that the performance has a larger variation among different combinations. The best average MAP performance of all datasets is under combination \textcircled{8}. The best performance on different datasets is under different combinations, which may be due to the data differences among datasets.

\subsubsection{The experimental analysis on the components of our framework}
\hspace*{0.5cm}Our framework proposed in this paper includes two stages, i.e. the kernel stage and quantization stage. The kernel stage in our framework is to fuse multiple views in a common space, and the quantization stage aims to learn the discriminative hash codes. To evaluate the importance of each stage, we perform the ablation experiments on WiKi, MMED, MIRFlickr, and NUS-WIDE respectively. Firstly, if we ignore the kernel stage and just use the quantization process, it means the input of our framework is a single view. As illustrated in Fig. 9, the performance of the 2nd-order statistics feature is superior to the 1st-order statistics feature and Histogram feature but is inferior to multi-view features. The MAP results of when using multi-view features exhibit improvements of  8.35\%, 3.18\%, 3.47\% over the second-order statistics feature.\\
\hspace*{0.5cm}Secondly, we record the retrieval results of MFDH when using arbitrary two views in the kernel stage to further validate how much each view contributes to the final retrieval result. As illustrated in Table 11, we can observe that MFDH using multi-view features achieves the best retrieval accuracy compared with the cases which just use two views. By analyzing the results in Table 11, we can easily find that the order is the 2nd-order statistics feature, 1st-order statistics feature and Histogram feature according to the contribution to final cross-modal retrieval result.   \\
\hspace*{0.5cm}By above analysis, we can conclude the following points: (1) Each view feature contributes to the final retrieval results; (2) The contribution of second-order feature is greater than others; (3) multiple views can promote jointly hashing learning and improve the retrieval accuracy.

\begin{figure*}[htbp]
\subfigure[WiKi]{
\includegraphics[width=.5\textwidth]{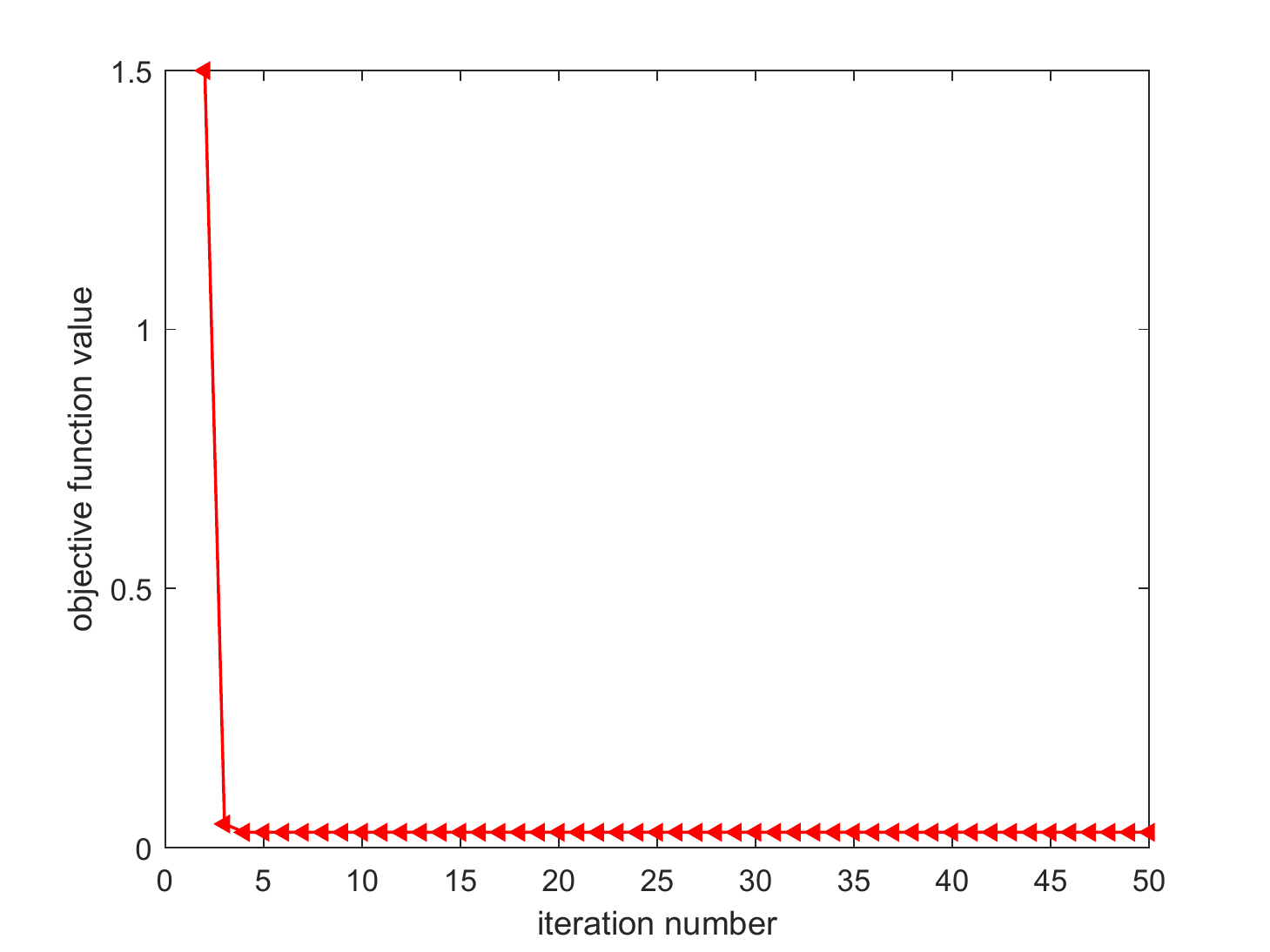}}
\subfigure[MMED]{
\includegraphics[width=.5\textwidth]{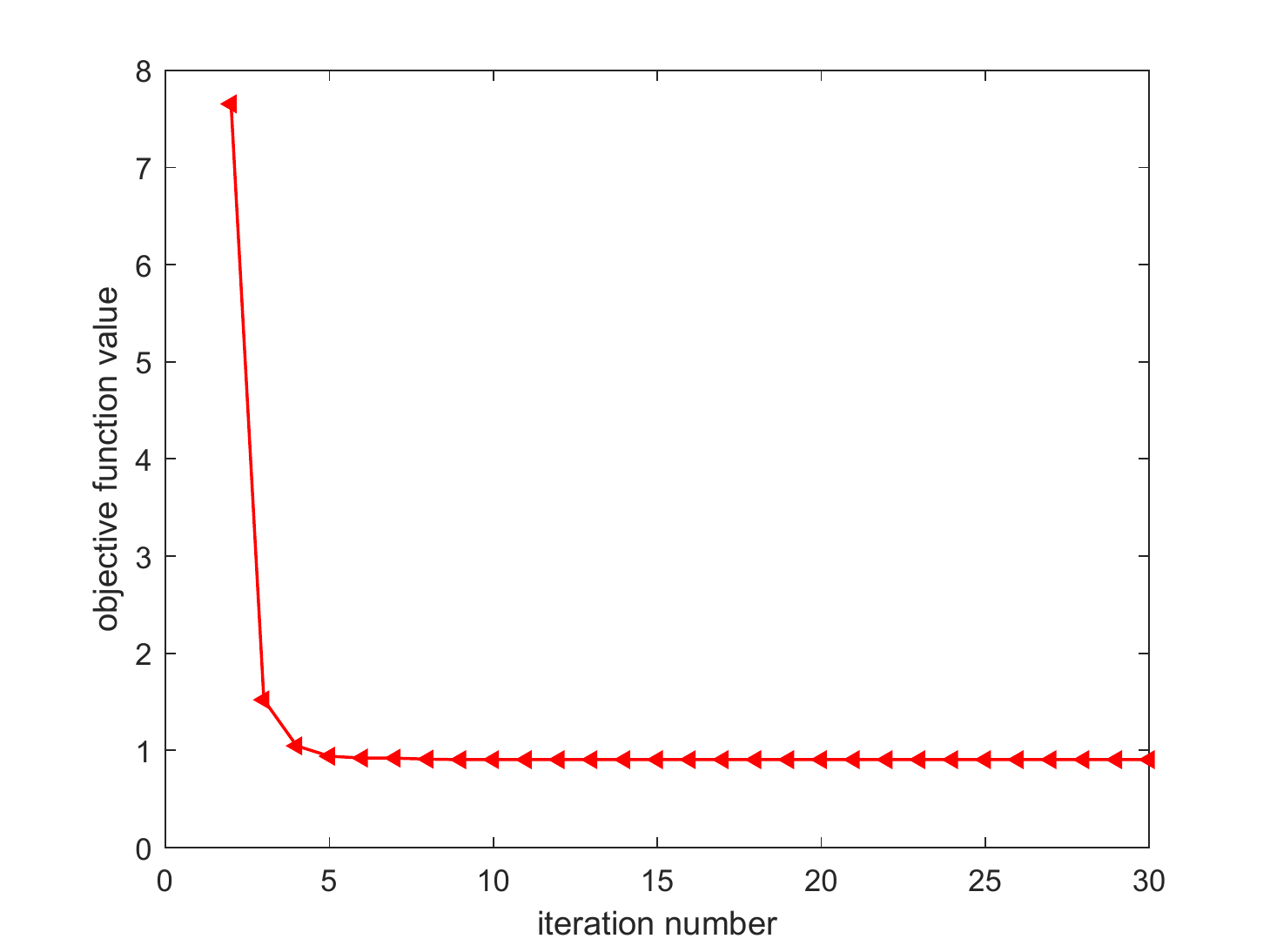}}
\subfigure[MIRFlickr]{
\includegraphics[width=.5\textwidth]{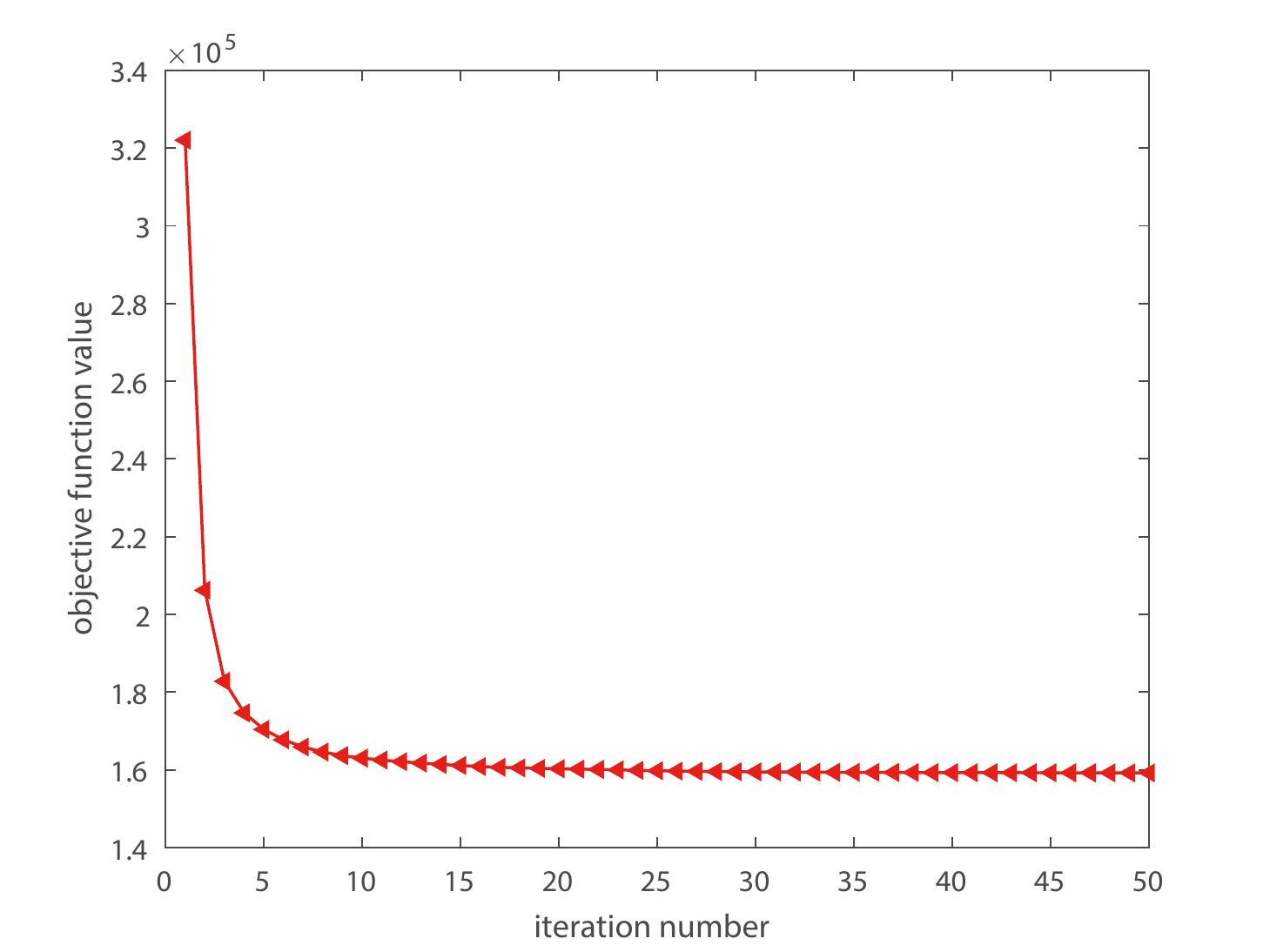}}
\subfigure[NUS-WIDE]{
%\centering
\includegraphics[width=.5\textwidth]{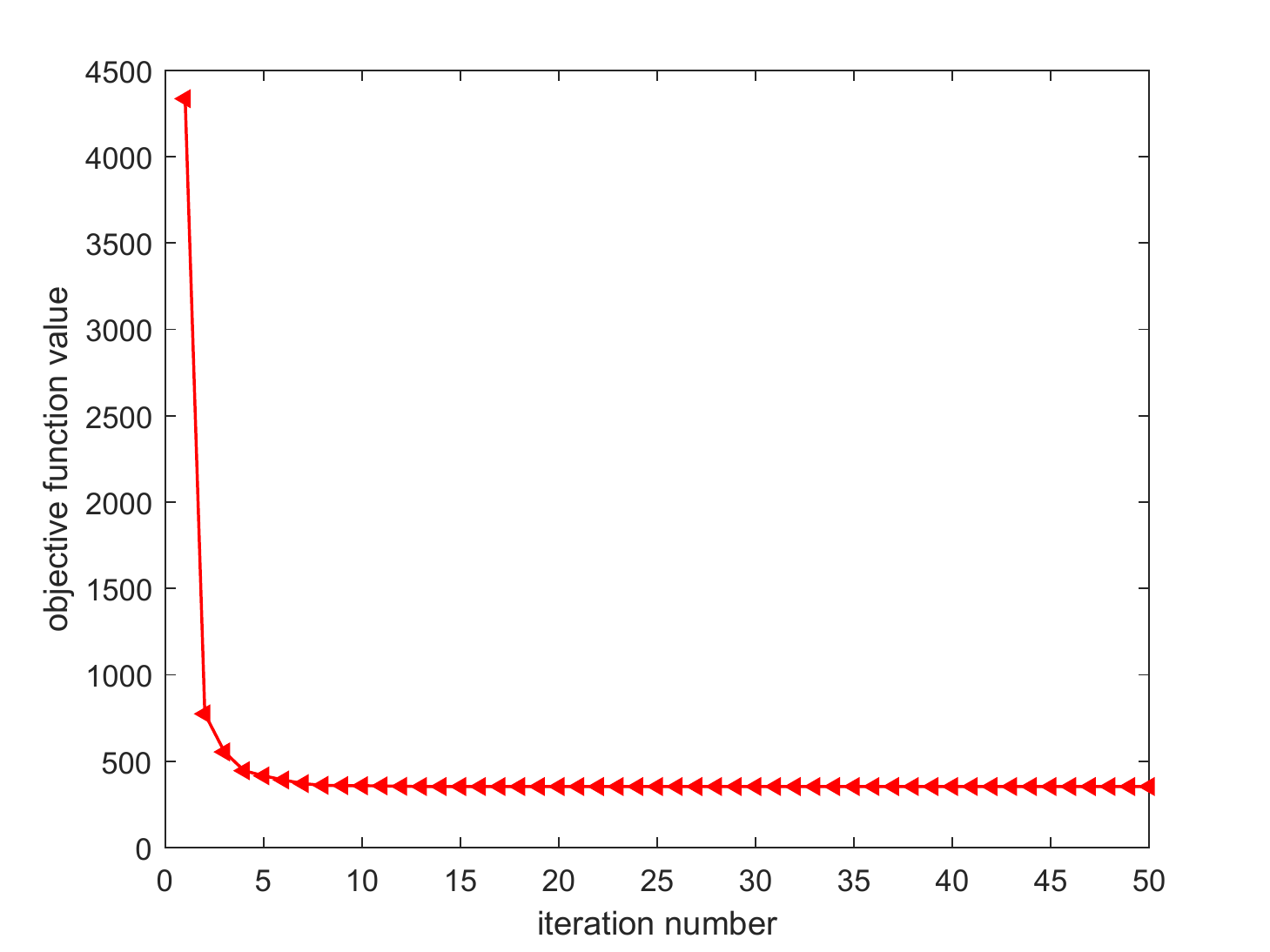}}
\caption{The convergence curves of Algorithm 1 on WiKi (a), MMED (b),  MIRFlickr (c), and NUS-WIDE (d).}
\end{figure*}

\subsubsection{The experimental analysis on the quantization stage of our framework}
\hspace*{0.5cm}Some experiments are conducted on four datasets to validate the contribution of each term in Equation 7. In Equation 7, the last term is classified into the first term. $\alpha$ and $\beta$ are two penalty parameters that control the weight of the second term and the third term respectively. MFDH($\alpha=0$) indicates that the model does not consider the role of the second term.  MFDH($\beta=0$) means the third item is discarded. MFDH($\alpha,\beta$) implies that we ignore the first item of the object function (7). As shown in Table 12, we have the following observations: (1) three terms work together to impact the cross-modal retrieval task, such as I2T and T2I ; (2) the first term and the third term have larger effect on Text query Text (T2T) task; (3) the first term and the second term contribute more to Image quey Image(I2I) task.

\subsection{Computational Complexity Analysis}
\hspace*{0.5cm}The computational complexity of MFDH contains two parts: computing kernelized features and the optimization procedures. Suppose that $d_I$ and $d_T$ represent the dimension of the descriptors for Image and Text respectively. The cost of generating kernelized features is $ \mathcal{O}(n\max(d_I,d_T)^3)$, where $n$ is the number of the training set. The optimization procedures involving the matrix multiplication requires $ \mathcal{O}((d^2L+L^2c+dcL)nT)$ to solve the optimization problem of Equation (7), where $d=\sum^2_{r=0}d_r$ and $T$ is the number of iterations. The total complexity of MFDH is $ \mathcal{O}(n\max(d_I,d_T)^3+(d^2L+L^2c+dcL)nT)$, where $L$ is the length of hash codes. The computational complexity is linear to the size of the training set, which is scalable to large-scale datasets.

\subsection{Convergence Analysis}
\hspace*{0.5cm}The optimization procedure of MFDH adopts the alternative iteration method shown in Algorithm 1. We show the convergence of Algorithm 1 by setting the length of the hashing code to 16 bits. For the other length of hashing codes, the convergence is similar. As shown in Fig.10, we can observe that the rate of convergence is very fast on all four datasets. Specifically, the algorithm converges within 10 iterations on WiKi and MMED, and within 30 iterations on MIRFlickr and NUS-WIDE. However, the theoretical convergence has not been found yet.\\

\section{CONCLUSION}
\hspace*{0.5cm}In this paper, we propose a new algorithm (MFDH) integrating the classifier learning and the subspace learning into a joint framework to learn discriminative compact hash codes for multiple retrieval tasks. MFDH mines the rich feature information of multimodal data in the form of multi-view features, and the semantic label information is exploited to learn the discriminative hashing functions. The proposed MFDH is evaluated on four multimodal datasets and the role of each component of our framework is explored experimentally. Extensive experimental results demonstrate that MFDH achieves significant performance compared with the state-of-the-art methods in terms of both accuracy and scalability. Future work will investigate the theoretical convergence of MFDH and consider the use of multiple kernel extension that integrates the feedback from the second stage to the first one in our framework.

%\section{ACKNOWLEDGMENT}
%This work is supported by the National Natural Science Foundation of China(Grant No.61373055,61672265), UK EPSRC Grant EP/N007743/1, MURI/EPSRC/dstl Grant EP/R018456/1, and the 111 Project of Ministry of Education of China (Grant No. B12018).

%\begin{acknowledgements}
%\end{acknowledgements}




\end{document}